\documentclass[nohyperref]{article}
\usepackage{microtype}
\usepackage{graphicx}
\usepackage{subfigure}

\usepackage{booktabs} 
\usepackage[accepted]{icml2022}

\usepackage{amsmath}
\usepackage{amssymb}
\usepackage{bbm}

\usepackage{caption}

\newtheorem{theorem}{Theorem}

\ifdefined\QED
\else
\newcommand{\QED}{\hfill \ensuremath{\Box}}
\fi

\icmltitlerunning{Efficient Computation of Higher-order Subgraph Attribution via Message Passing}

\begin{document}

\twocolumn[
\icmltitle{Efficient Higher-order Subgraph Attribution via Message Passing}

\begin{icmlauthorlist}

\icmlauthor{Ping Xiong}{tu}
\icmlauthor{Thomas Schnake }{tu,bf}
\icmlauthor{Gr\'{e}goire Montavon}{tu,bf}
\icmlauthor{Klaus-Robert M\"{u}ller}{tu,bf,korea,saar}
\icmlauthor{Shinichi Nakajima}{tu,bf,riken}

\end{icmlauthorlist}

\icmlaffiliation{tu}{Technische Universit\"{a}t Berlin (TU Berlin)}
\icmlaffiliation{bf}{BIFOLD -- Berlin  Institute  for  the Foundations  of  Learning  and  Data}
\icmlaffiliation{korea}{Department of Artificial Intelligence, Korea University, Seoul 136-713, Korea}
\icmlaffiliation{saar}{Max Planck Institut für Informatik, 66123 Saarbrücken, Germany}
\icmlaffiliation{riken}{RIKEN Center for AIP, Japan}

\icmlcorrespondingauthor{Shinichi Nakajima}{nakajima@tu-berlin.de}
\icmlkeywords{Graph Neural Networks, Explanation, Message Passing}

\vskip 0.3in
]

\printAffiliationsAndNotice{} 

\begin{abstract}
Explaining graph neural networks (GNNs) has become more and more important  recently. Higher-order interpretation schemes, such as GNN-LRP (layer-wise relevance propagation for GNN), emerged as powerful tools for unraveling how different features interact thereby contributing to  explaining GNNs.
GNN-LRP gives a relevance attribution of walks between nodes at each layer, and the subgraph attribution is expressed as a sum over exponentially many such walks. In this work, we demonstrate that such exponential complexity can be avoided.  In particular, we propose novel algorithms
that enable to attribute subgraphs with GNN-LRP in linear-time (w.r.t. the network depth). 
Our algorithms are derived via message passing techniques that make use of the distributive property, thereby
directly computing quantities
for higher-order explanations.
We further adapt our efficient algorithms to compute
a generalization of subgraph attributions that also takes into account the neighboring graph features.
Experimental results show the significant acceleration of the proposed algorithms and demonstrate the high usefulness and scalability of our novel generalized subgraph attribution method.
\end{abstract}

\section{Introduction}
\begin{figure}
    \centering
     \includegraphics[width=1.0\linewidth]{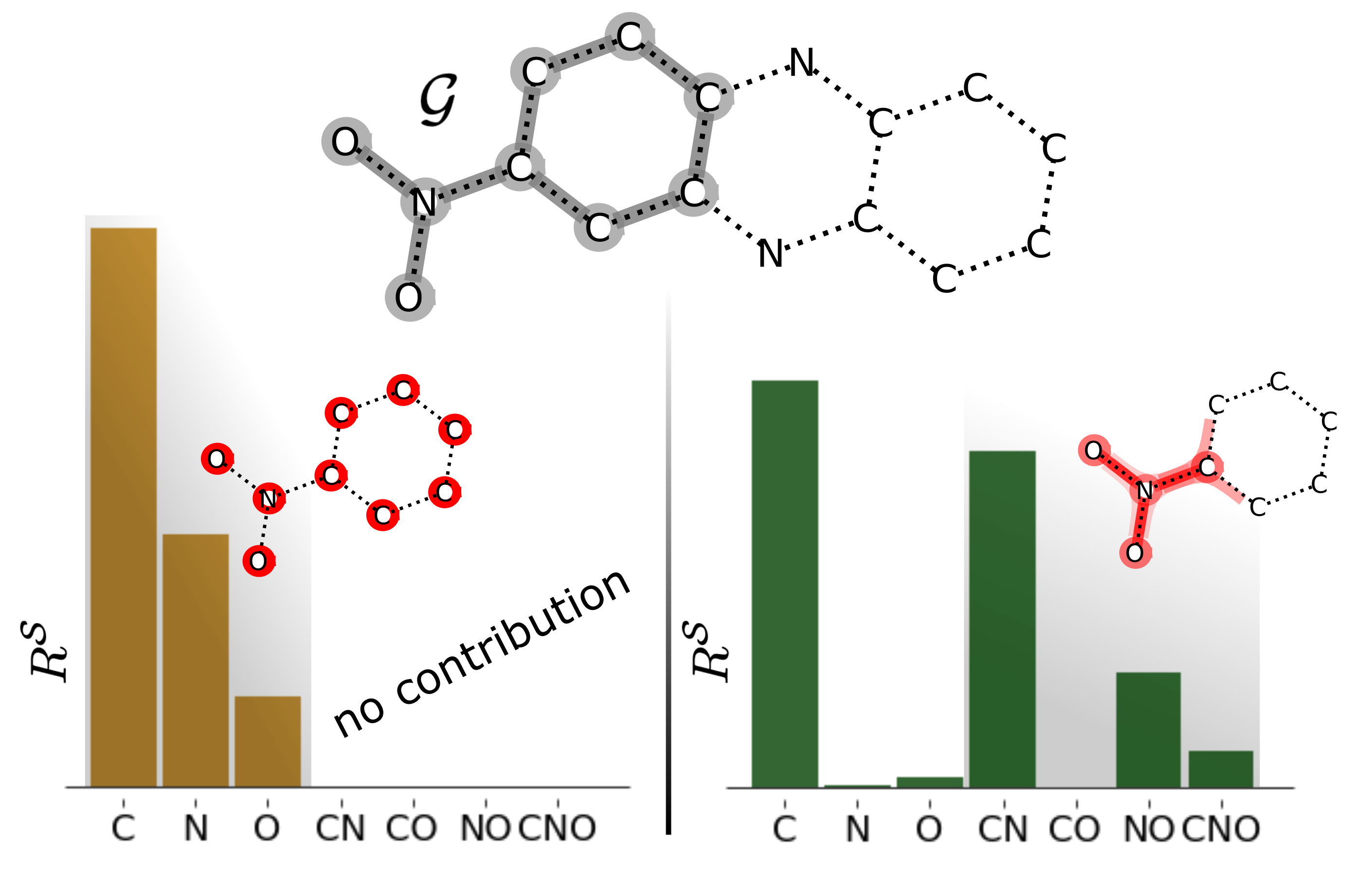}
    \vspace{-3mm}
    \caption{Visualisation for the joint contribution of atom subsets when predicting the mutagenicity of a molecule. The gray sub-molecule $\mathcal{G}$ is a strong indicator for the mutagenicity of the full molecule. The orange and green bars show the lower- and higher-order relevance scores for the subset composed of all atoms which have the denoted atomic number, respectively. The heat maps in the sub-molecules, close to the orange and green bars, show the relevance scores of  lower- and higher-order attribution methods, respectively. 
}
    \vspace{-1mm}
    \label{fig:comp_holo_mol}
\end{figure}

 \begin{figure*}
     \centering
     \includegraphics[width=\textwidth]{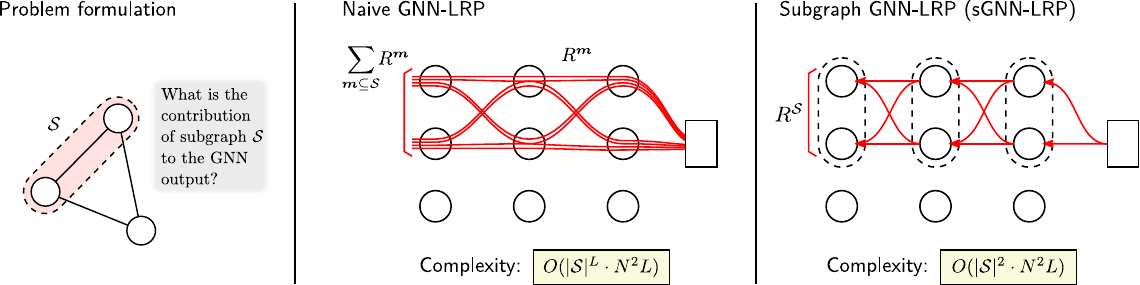}
     \vspace{-2mm}
     \caption{Computation of the subgraph relevance (left).  A naive implementation of GNN-LRP computes the sum of the relevances over exponentially many walks (middle), while our proposed subgraph GNN-LRP aggregates the contributions \emph{layer-wise} (right), allowing linear time computation with respect to the network depth $L$.
     }
     \label{fig:conceptual_figure}
 \end{figure*}

In recent years, there has been an increasing interest in Graph Neural Networks (GNNs) because of their ability to incorporate the intrinsic structure of  data and their state-of-the-art performance on graph-structured data, e.g., social networks \cite{chen2018fastgcn, hamilton2017inductive, DBLP:conf/iclr/KipfW17} and molecules \cite{schutt2018schnet,domingue2019evolution}. The learning tasks on graphs include node, edge or graph classification, link prediction and others \cite{wu2020comprehensive, hu2020ogb}. However, since the prediction strategy of a GNN is in general not comprehensible for humans, GNN models are still treated as black-boxes, which prevents applications in some crucial areas where trustworthiness or safety is required. In the recent literature, various methods for explaining GNNs have been developed \cite{yuan2020explainability}. 
Methods like GNNExplainer \cite{ying2019gnnexplainer}, PGExplainer \cite{luo2020parameterized} and PGM-explainer \cite{vu2020pgm}
allow importance analysis of nodes and edges within the input data sample. GNN-LRP (layer-wise relevance propagation for GNN; \citet{schnake2020higher}) aims at explaining GNNs at the level of \emph{walks}, which reflect the practically relevant higher-order interactions of features. To obtain such walk relevances,  higher-order deep Taylor decomposition is applied to a GNN, from which we get independent feature components that only depend on bag-of-edges. These bag-of-edge components are then recovered by backward messages from one node representation to the next within the GNN interaction layers, and provide the relevance resolution of walks \citep{schnake2020higher}.

Higher-order interpretation methods can give important insights into the prediction strategies employed by neural networks. In Figure \ref{fig:comp_holo_mol} we see how lower- and higher- order interpretation methods assign contributions to 
different subsets of atoms when predicting the mutagenicity of molecules \cite{doi:10.1021/jm040835a}. Using the higher order scheme, we find that the $\text{NO}_2$ group combined with the neighboring aromatic carbon ring (highlighted in gray in Figure 1), is a strong indicator for mutagenicity. This finding is also supported by experimental studies \cite{Jeroen205derivmutag}. The lower order attribution method is unable to capture the joint contributions of atoms, and therefore have to rely on single atom contributions exclusively, inappropriately assuming that the individual atoms contribute to the mutagenicity of the molecule. On the other hand, the attribution scheme that takes into account the higher-order structure of the model reflects how sets of different atoms, particularly the combination of C and N, or N and O atoms, have a strong joint contribution to the prediction task. This aligns with chemical intuition in general, since the property of mutagenicity is not due to the effect of single atoms but arises from the higher-order interactions present in the highlighted functional group. More details can be found in Appendix~\ref{app:fig1}.

Although GNN-LRP 
has shown to be
highly effective 
in interpreting GNNs with respect to feature interactions, its computation was so far limited to relatively small graphs and shallow networks, due to the exponential complexity with respect to the network depth.
We would like to note that this complexity issue is not specific to GNN-LRP but rather it is present for general higher-order feature attribution methods beyond additive or linear explanation  \cite{lund2017unified,samek2021explaining}. This is because $L$-th order interpretation methods for $N$ input features need to take $N^L$ different feature combinations into account.
In the task of extracting the joint relevance of a collection of graph features, namely the relevance of a subgraph in the input data point \cite{schnake2020higher, yuan2021explainability}, the incorporation of higher-order feature attribution is essential, yet without finding an efficient way of computation that remedies the exponential complexity, it is unfeasible for the general case even with moderately large $L$. 

In this work,
we propose a novel propagation rule, called subgraph GNN-LRP (sGNN-LRP), that directly computes the relevance of a subgraph in a single backpropagation pass.
Comparing with a naive application of GNN-LRP that sums up walk relevances, the computational complexity reduces from exponential to linear with respect to the network depth $L$ (see Figure~\ref{fig:conceptual_figure}).
The forward-hook trick \cite{schnake2020higher, samek2021explaining} allows 
a simple, fast, and less memory intensive implementation of sGNN-LRP.

A novel aspect of this work also exists in the way of developing the new propagation rule: sGNN-LRP is derived as a sum-product message passing algorithm, a.k.a., belief propagation \cite{bishop2006patternchp8, DBLP:conf/aaai/Pearl82}, to compute an explicitly defined target quantity---the sum of relevances of all walks that stay within a given subgraph.  We explain why message passing is applicable to the relevance computation by pointing out its mathematical similarity to the marginal probability computation of a Markov chain process, and discuss its generality by deriving existing LRP rules as message passing algorithms.

{ 
\def\arraystretch{1.3}\tabcolsep=6pt

\begin{table}[t]
    \caption{Notation.}
    \label{tab:notation_table}
\vspace{-1mm}
    \begin{center}
        \begin{small}
                \begin{tabular}{|c||p{0.31\textwidth}|}
                    \hline
                    $h, \mathbf h, \mathbf H$, $H_{m,m'}$ & scalar, vector, matrix, matrix entry \\
                    $ m_{l:l'} $ & partial vector with indices $(l, \dots, l') $ \\ 
                    \hline
                    $\mathcal G$ and $ \mathcal S$ & graph and subgraph \\
                    $\mathbf m$ and $ \mathbf n$ & sequence of nodes and neurons \\
                    $m$, $m_l$ & integers for node identifications \\
                    $n$, $n_l$ & integers for neuron identifications \\
                    $R$, $\mathbf r$ & relevance \\
                    $\breve{\mathbf r}$ & propagated relevance, message, or belief \\
                    $\mathbf T$ & propagation matrix\\                   
                    \hline
                \end{tabular}
        \end{small}
    \end{center}
\end{table}

}

The message passing framework allows us to easily adapt the propagation rule to another target quantity.  We demonstrate this benefit by deriving a variant of sGNN-LRP for a generalized definition of the subgraph relevance, which takes into account the walks outside the subgraph with discounted contributions according to how many times the walk steps out of the subgraph.
Our experiments show that our generalized subgraph relevance 
quantitatively improves the explanation of GNNs in terms of node-ordering performance.

\section{Background and Related Work}

\subsection{Graph Neural Networks}

Graph Neural Networks (GNNs) \cite{DBLP:journals/tnn/ScarselliGTHM09, wu2020comprehensive} is a class of neural networks that receive a graph as an input. In a GNN, node embeddings 
are learned in multiple \textit{interaction blocks},
where the interaction is defined by the given graph.
In most GNN architectures,
the interaction block
can be divided into \textit{aggregate} and \textit{combine} steps \citep{gilmer2017neural}, which can be expressed by
\begin{equation}
    \begin{aligned}
        \text{Aggregate: } \mathbf Z^{(l)} &=\mathbf \Lambda \mathbf H^{(l-1)}, \\
        \text{Combine: } \mathbf H^{(l)} &= \textstyle \mathcal C^{(l)}\left(\mathbf Z^{(l)}\right).
    \end{aligned}
    \label{eq:GNN_agg_comb}
\end{equation}
Here $\mathbf H^{(l)} \in \mathbb R^{M\times N^{(l)}}$ denotes the feature (activation) matrix of the $l$-th layer, which consists of the $N^{(l)}$-dimensional feature vectors for all $M$ nodes (notation is summarized in Table \ref{tab:notation_table}).
In the aggregate step, the features $\mathbf H^{(l-1)}$ from the last layer 
are aggregated using a modified (e.g., normalized with self-loops) adjacency matrix $\mathbf \Lambda \in \mathbb  R^{M\times M}$ 
and stored in $\mathbf Z^{(l)} \in \mathbb  R^{M\times N^{(l-1)}}$.
In the combine step, a non-linear function $\mathcal C^{(l)}: \mathbb R^{N^{(l-1)}} \mapsto \mathbb R^{N^{(l)}}$ is applied to 
$\mathbf Z^{(l)}$ column-wise,
to transform $\mathbf Z^{(l)}$ into the new node features $\mathbf H^{(l)}$ for this layer. 
For the combine function, 
common choices are a one-layer perceptron 
$\mathcal C^{(l)}\left(\mathbf Z\right)= \sigma(\mathbf Z \mathbf W^{(l)})$ with trainable weights $\mathbf W^{(l)} \in \mathbb R^{N^{(l-1)} \times N^{(l)}}$ and a non-linear activation  $\sigma(\cdot)$
as in 
 Graph Convolution Network (GCN) \cite{DBLP:conf/iclr/KipfW17}, 
and a multilayer perceptron (MLP) as in 
Graph Isomorphism Network (GIN) \cite{xu2018powerful}. %
After the node feature of the last layer is computed, a read-out function (e.g., average or maximum over all nodes, followed by an MLP) is applied to generate the required graph-level predictions, depending on
the prediction task. 

\subsection{Explainability for Graph Neural Networks}

In recent years multiple explainability methods for GNNs have been proposed. 
GNNExplainer \cite{ying2019gnnexplainer} and PGExplainer \cite{luo2020parameterized} explain GNNs by finding masks that maximize the mutual information between the predictions of the original graph and a masked graph. GNNExplainer learns soft masks for edges or node features, while PGExplainer trains a parametric predictor to determine if an edge should be masked out.  The predicted mask is an approximate discrete mask and is known to alleviate the `introduced evidence' problem that soft mask faces. Such masks can be used to extract the most important subgraphs, for example, we can identify the subgraph consisting of all nodes with soft mask values above a threshold as the most important subgraph.
PGM-Explainer \cite{vu2020pgm} trains a well explainable probabilistic graphical model (PGM) as a surrogate method of the GNN, and then substitute the explanation of the GNN with the explanation of the PGM. Unlike all the instance-level methods above, XGNN \cite{DBLP:conf/kdd/YuanTHJ20} is a model-level explainer, which generates a representative graph of every target class using reinforcement learning.

Most explainability techniques \cite{DBLP:conf/cvpr/PopeKRMH19, ying2019gnnexplainer, luo2020parameterized} for GNNs explains the model at the level of nodes, edges and node features, while a few of them, including SubgraphX \cite{yuan2021explainability} and GNN-LRP \cite{schnake2020higher}, analyze the relevance of subgraphs as higher-order features. 
SubgraphX identifies the most important subgraphs
based on the Shapley value \cite{lund2017unified}
with Monte-Carlo Tree Search (MCTS).
The Shapley value is computed by perturbing the input graph and comparing the change of the model output.
GNN-LRP is an LRP-based method, which scores bag-of-edges by decomposing and backpropagating the output to the input layer.
Since our proposed algorithm is based on GNN-LRP,  %
we give its detailed description in the next subsection.

Unlike the above methods, GNES \cite{DBLP:conf/icdm/GaoSBYH021} provides a general framework that trains the GNN model and optimizes the explanation model simultaneously with regularizations, so that the explanation is reasonable and stable. 
GNES can handle many attribution methods including Gradient-based, Grad-CAM and LRP.

\subsection{GNN-LRP}

GNN-LRP \cite{schnake2020higher} 
aims to explain the prediction strategy of GNNs with respect to higher-order feature interactions, 
by tracking the information that were passed through the internal dependency graph for making a prediction.
The basic unit of explanation is 
therefore the relevance of a \emph{walk}, which is defined as an ordered sequence of nodes connected from layer to layer.
Assume that the whole graph $\mathcal G$
consists of $M$ nodes.  Then, a walk can be denoted by $\mathbf m \in \mathbb{M} $ with $\mathbb{M}  = \{1, \ldots, M\}^{L+1}$, meaning that the walk starts from the $m_0$-th node at the input layer, goes through the $m_l$-th node at the $l$-th layer, and reaches the $m_L$-th node in the last layer.
We also denote a partial walk by $m_{l:l'}$ for $0 \leq l < l'  \leq L$.
We identify a node and its index, and denote, e.g., by $m \in \mathcal{G}$ that the node $m$ is a member of $\mathcal{G}$.

The GNN-LRP rule for a simple GCN \cite{DBLP:conf/iclr/KipfW17} with the combine function $\mathcal C^{(l)}(\mathbf Z) = \text{ReLU} \left(\mathbf Z \mathbf W^{(l)}\right)$ in Eq.\eqref{eq:GNN_agg_comb} is given as
\begin{align}
  \breve{  \mathbf  r}^{(l, m_l)} = \mathbf T^{l, m_l, m_{l+1}}   \breve{  \mathbf  r}^{(l+1, m_{l+1})}.
    \label{eq:GCN_GNN_LRP_rule}
    \end{align}
    Here 
    $\breve{\mathbf{r}}^{(l, m_l)} \in \mathbb R^{N^{(l)}}$ is the \emph{propagated relevance} at the node $m_l$ in the $l$-th layer,
    and $\mathbf T^{l, m, m'} \in \mathbb R^{N^{(l)} \times N^{(l+1)}}$ is 
    the propagation matrix whose entries are given as
\begin{align}
 T_{n, n'}^{l, m, m'} = \textstyle
\frac{\Lambda_{m, m'} H_{m, n}^{(l)} W^{(l)\uparrow}_{n, n'} }{\sum_{m'' ,n''} {\Lambda_{m'', m'} H_{m'', n''}^{(l)} W^{(l)\uparrow}_{n'', n' } }}.
\notag
\end{align}
Here
$ \mathbf{W}^{\uparrow}$ is a modified weight parameter depending on the choice of LRP rules \cite{bach2015pixel,montavon2018methods,samek2021explaining}, e.g., 
$\mathbf{W}^{\uparrow} := \mathbf{W} + \gamma\cdot \max(0, \mathbf{W} )$ for the LRP-$\gamma$ rule with $\gamma \geq 0 $,
where the max operator applies entry-wise.
Note that we mostly use subscripts to specify the entry of a matrix or vector, while superscripts
for distinguishing different matrices or vectors.
Note also that the propagated relevance $\breve{\mathbf{r}}^{(l, m_l)}$ does not denote a  particular quantity but a variable that depends on the propagation rule. 

The GNN-LRP rule depends on the network structure, and 
we refer to \citet{schnake2020higher} for the GNN-LRP rules for other GNN variants.

\section{Efficient Computation of Subgraph Attribution }
\label{sec:MPAlgorithmProposal}

Understanding the relevance contribution of subgraphs in the input graph to the model prediction is a key challenge when explaining models on graphs \cite{yuan2021explainability, luo2020parameterized, schnake2020higher}. As a higher-order interpretation method, 
\citet{schnake2020higher} proposed a definition of subgraph relevance as the sum over relevance scores of all walks inside the subgraph $\mathcal S \subseteq \mathcal G$, i.e.,
\begin{align}
    R^{\mathcal S} = \textstyle \sum_{ \mathbf m \subseteq \mathcal S} R^{ \mathbf m}
    \label{eq:subgraph_rel_def_1}.
    \end{align}
Here, with slight abuse of notation, we denote by $\mathbf m \subseteq \mathcal S$ that the walk $\mathbf m$ stays inside $\mathcal S$, i.e., $m_l \in \mathcal S $ for all $l=0, \ldots, L$.
Unfortunately, performing the sum in Eq.\eqref{eq:subgraph_rel_def_1}, 
which we call a \emph{naive} application of GNN-LRP for subgraph attribution (Naive GNN-LRP), over exponentially many $\sim \mathcal{O}(|\mathcal{S}|^L)$ walks is limited to small subgraphs (nodes or edges) for state-of-the-art GNNs as they are typically deep.

\begin{figure*}
    \centering
    \subfigure[Input graph]{ 
    \begin{minipage}{4cm}
    \centering
    \includegraphics[scale=0.7]{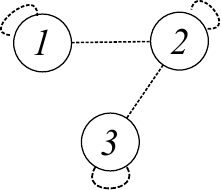} 
    \end{minipage}
    }
    \subfigure[GNN-LRP for a walk]{ 
    \begin{minipage}{6cm}
    \centering 
    \includegraphics[scale=0.41]{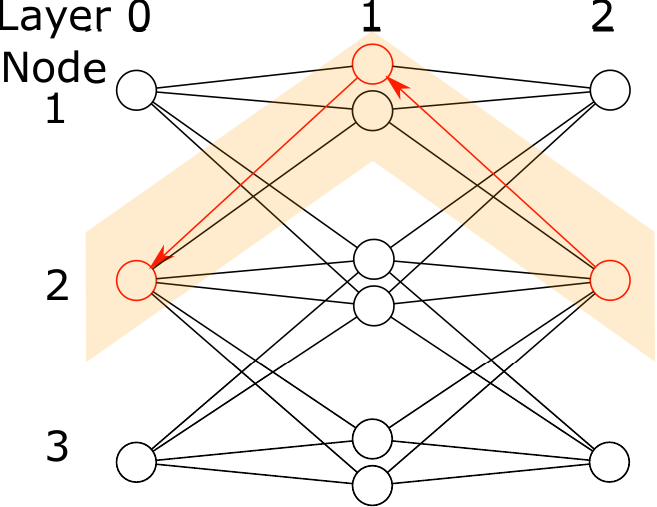}
    \end{minipage}
    }
    \subfigure[Markov Chain]{ 
    \begin{minipage}{6cm}
    \centering 
    \includegraphics[scale=0.45]{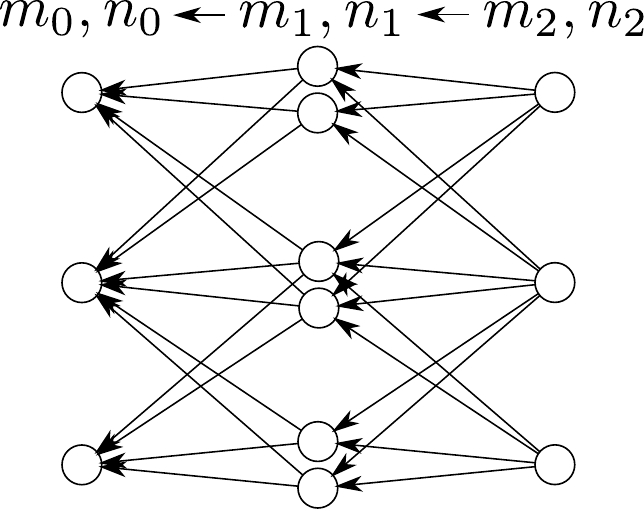}
    \end{minipage}
    }
    \vspace{-2mm}
    \caption{LRP as message passing. (a) An example 
    graph consisting of 3 nodes (with self-connections).
    (b)
    Unfolded feed-forward network of a 2-layer GCN (with the node feature dimension in each layer being $N^{(0:2)} = [1, 2, 1]$).
    A walk ($\mathbf m = [2,1,2]$) is marked with yellow background, and 
    a neuron-level walk $(\mathbf m = [2,1,2], \mathbf n = [1,1,1]$) is marked as red arrows. (c) A Markov chain process of which the joint distribution has the same decomposable structure as the relevance of a neuron-level walk.  This implies that message passing techniques for computing various marginal probabilities of Markov chain can be used for computing the sum of relevances over various sets of walks.
    }
    \vspace{-3mm}
    \label{fig:conceptual_figure_markov}
\end{figure*}

However, 
we will show in Section~\ref{sec:LRPasMP} that
there is a much more efficient way of achieving subgraph attribution: the exponential sum in Eq.\eqref{eq:subgraph_rel_def_1} can be computed by
$
R^{\mathcal S} =\textstyle \sum_{m_0 \in \mathcal{S}} \sum_{n=1}^{N^{(0)}}   \breve{r}_n^{(0, m_0)} 
\label{eq:GCN_GNN_LRP_rule_2.FinalResult}$
after applying a single pass of our novel subgraph GNN-LRP (sGNN-LRP) rule
\begin{align}
   \breve{\mathbf{r}}^{(l, m_l)} = \textstyle \sum_{m_{l+1} \in \mathcal S} \mathbf T^{l, m_l, m_{l+1}}   \breve{\mathbf{r}}^{(l+1, m_{l+1})}
    \label{eq:GCN_GNN_LRP_rule_2}
\end{align}
for $l = L-1, \ldots, 0$.
Since a single backward pass directly yields 
the subgraph attribution,
 the computational advantage is evident and massive,
 as shown graphically in Figure~\ref{fig:conceptual_figure}. 
Moreover, applying
the \emph{forward-hook} trick
\cite{schnake2020higher, samek2021explaining}
to sGNN-LRP gives
a simpler, faster, and less memory intensive implementation
than directly implementing the sGNN-LRP rule (see Section~\ref{sec:ForwardHook}).

Our novel sGNN-LRP rule \eqref{eq:GCN_GNN_LRP_rule_2} only differs from the GNN-LRP rule \eqref{eq:GCN_GNN_LRP_rule} for computing the walk relevance, by introducing an additional summation that pools the propagated relevances over all nodes belonging to the subgraph $\mathcal{S}$. 
However, we emphasize that we obtained this new rule by a novel, systematic procedure: we explicitly define the target quantity to be computed, and then derive a propagation rule as a message passing algorithm.
We detail this procedure and its generality in Section~\ref{sec:LRPasMP}.

\section{LRP as Message Passing}
\label{sec:LRPasMP}

In this section, we first point out the similarity between the relevance computation and the marginal probability computation of a Markov chain, which implies the applicability of \textit{sum-product}  (a.k.a. belief propagation) algorithm \cite{bishop2006patternchp8, DBLP:conf/aaai/Pearl82}, a message passing algorithm for marginalizing over random variables.
Then, we apply the sum-product decomposition to the target quantity \eqref{eq:subgraph_rel_def_1},
and derive our sGNN-LRP rule \eqref{eq:GCN_GNN_LRP_rule_2}.
We also discuss the generality of our approach, and derive existing LRP rules.
This novel procedure allows us to systematically derive new propagation rules by defining or modifying the target quantity, which will be further demonstrated in Section~\ref{sec:GeneralizedSubgraphAttribution}.

Let us define the relevance of a partial walk $m_{l:l'}$
 as the sum of relevances of all walks going through the specified nodes from the $l$-th to the $l'$-th layers, i.e.,
\begin{align*}
    R^{m_{l:l'}} = \textstyle \sum_{\substack{\mathbf{m}' \in \mathbb{M}: m'_{l:l'} \, = \, m_{l:l'} } }  R^{\mathbf{m}'},
\end{align*}
and its neuron-level counterpart $\mathbf{r}^{l, m_{l:l'}} \in \mathbb{R}^{N^{(l)}}$ 
   whose entry   $r_n^{l, m_{l: l'}}$  is the relevance of a partial walk limited to a particular neuron specified by $n$ at the $l$-th layer.
We denote by $m_{l: l'} \subseteq \mathcal S$ that the partial walk is within the subgraph, i.e., $m_{l''} \in \mathcal S$ for $l'' = l, \ldots, l'$,
and by $\mathbf r^{l, m_l, m_{l+1: L} \subseteq \mathcal S}$ the sum of relevances of the partial walks that go through the node $m_l$ at layer $l$ and any node in $\mathcal{S}$ at layers $l+1, \ldots, L$.

GNNs can be unfolded into a feed forward neural network (FFNN) (Fig.\ref{fig:conceptual_figure_markov}(b)).
Based on the unfolded network,
consider a virtual stochastic system where a particle exists at the $L$-th layer at time $t = 0$, stays at the $l$-th layer at time $t = L - l$, and arrives at the input layer at time $t = L$.  At each layer, the particle  stochastically chooses a particular neuron in a particular node.  Then, its trajectory can be denoted by $(\mathbf{m}, \mathbf{n})$, where $\mathbf{n} \in \mathbb{N} \equiv \{1, \ldots, \max_l (N^{(l)})\}^{L+1}$ specifies the choice of neuron at each layer.  Assume that the probability of the choice of node-neuron pair at the $l$-th layer only depends on the choice at the $(l+1)$-th layer.
Then,
the joint probability can be written as
\begin{align}
\textstyle
p(\mathbf{m}, \mathbf{n}) = \left(\prod_{l=0}^{L-1} p(m_l, n_l | m_{l+1}, n_{l+1}) \right) p(m_L, n_L) ,
\label{eq:JointDistributionMarkov}
\end{align}
which is a simple Markov chain (Fig.\ref{fig:conceptual_figure_markov} (c)).
If we formally assume that $p(m_l, n_l | m_{l+1}, n_{l+1}) =  T_{n_{l}, n_{l+1}}^{l, m_l, m_{l+1}} $
and $p(m_L, n_L) = r_{n_L}^{L, m_L} $, the joint distribution \eqref{eq:JointDistributionMarkov} coincides with the relevance of a \emph{neuron-level} walk, a walk 
specifying not only the node but also the neuron inside the node at each layer:
\begin{align}
R^{\mathbf{m}, \mathbf{n}} =\textstyle  \left(\prod_{l=0}^{L-1} T_{n_{l}, n_{l+1}}^{l, m_l, m_{l+1}} \right) 
r_{n_L}^{L, m_L} = p(\mathbf{m}, \mathbf{n}) .
\label{eq:NeuronlevelWalkRelevance}
\end{align}
Importantly, the relevance has the same decomposable structure as the joint distribution of the Markov chain.
Therefore, we can use the sum-product algorithm---which allows efficient computation for various marginal distributions of a Markov chain---for computing relevances that require summation over different sets of walks.

Since
the propagation matrices $\{\mathbf{T}^{l, m, m'}\} $ and the partial walk relevance $\mathbf{r}^{L, m_L} $ 
are not probabilities, they can have negative entries,
and \emph{not} necessarily normalized.  These differences do not affect the applicability of the sum-product decomposition, and we can derive LRP rules as message passing for any propagation matrices and any definition of relevance.  However, we restrict our theoretical analysis in this paper to the case where the propagation matrix is normalized, i.e., $\sum_{n, m} T_{n, n'}^{l, m, m'}=1 \; \forall n', m' $ for simplicity.  This allows us to precisely and concisely describe what quantity is carried by the propagated relevance $\breve{\mathbf r}^{(l, m_l)}$ (which corresponds to the message/belief in the terminology for message passing/belief propagation), and makes the derived LRP rules transparent.
For unnormalized propagation matrices, the messages are scaled by
layer-dependent constants, which are practically irrelevant and (if necessary) can be computed by another pass of messages.

Now let us derive our sGNN-LRP rule \eqref{eq:GCN_GNN_LRP_rule_2}.  
Setting Eq.\eqref{eq:subgraph_rel_def_1} as the target quantity,%
\footnote{
The subgraph relevance \eqref{eq:subgraph_rel_def_1}  corresponds to the marginal probability that a particle never steps out of the subgraph $\mathcal{S}$  in the virtual Makov chain process shown in Fig.\ref{fig:conceptual_figure_markov}(c).
}
we apply the sum-product decomposition, and obtain the following theorem 
(the proof is given in Appendix~\ref{sec:ProofSP}):
\begin{theorem} \label{thm:sumprod}
  Assume that the sGNN-LRP rule \eqref{eq:GCN_GNN_LRP_rule_2} is applied for $l = L-1, \ldots, 0$ with the initial message $\breve{\mathbf{r}}^{(L, m_L)} = \mathbf{r}^{L, m_L}$.  Then, $\breve{\mathbf{r}}^{(l, m_l)} = \mathbf{r}^{l, m_l, m_{l+1: L } \subseteq \mathcal S} \; \forall l \in \{0, \ldots, L\} $, and 
  $R^{\mathcal S} = \sum_{m_0 \in \mathcal{S}} \sum_{n=1}^{N^{(0)}}  r_n^{0, m_0, m_{1: L } \subseteq \mathcal S} $.

\end{theorem}

Our novel procedure---deriving LRP rules as message passing for computing explicitly defined target quantities---is general, and one can derive many existing LRP rules by defining the corresponding target values,
as summarized in
Table~\ref{table:GeneralLRPTable}
(see Appendices~\ref{sec:LRPasSumProduct} and \ref{sec:ProofSPGNNLRP} for derivations, and Appendix \ref{appendix:schnake-notation} for the same rules using the notation of \citet{schnake2020higher}).
This procedure allows systematic derivation of propagation, and will facilitate future development of LRP methods.

\begin{table*}[t]
\caption{Target quantities and the corresponding propagation rules derived as message passing. The propagation matrices $\{  \mathbf  T^{l} \in \mathbb R^{N^{(l)} \times N^{(l+1)}}\}$ can be arbitrary, and therefore this table applies for all ($\epsilon, \gamma, \alpha\beta$, etc.) propagation matrices. 
}
\vspace{-4mm}
\center
\begin{tabular}{lll}
\toprule
& Target quantity & Propagation rule\\\midrule
LRP for general FFNN & $ \mathbf r^{0} =  (    \prod_{l=0}^{L-1}
   \mathbf  T^{l}
    )
 \mathbf r^{L}$ & $ \breve{ \mathbf r}^{(l)} =  \mathbf T^{l} \breve{  \mathbf r}^{(l+1)} $\\
GNN-LRP & $R^{\mathbf m}$ & $\breve{\mathbf r}^{(l, m_{l})} = \mathbf T^{l, m_l, m_{l+1}}     \breve{\mathbf  r}^{(l+1, m_{l+1})}$ \\
sGNN-LRP &$R^{\mathcal S} = \sum_{\mathbf m \subseteq \mathcal S}
            R^{\mathbf m}$ & $\breve{ \mathbf  r}^{(l, m_{l})} = \textstyle \sum_{m_{l+1} \in \mathcal S} \mathbf T^{l, m_l, m_{l+1}}   \breve{  \mathbf  r}^{(l+1, m_{l+1})}$ \\
Generalized sGNN-LRP  & $ \widetilde{R}_\alpha^{\; \mathcal S} =  \sum_{\mathbf m \subseteq \mathcal G}
         \alpha^{ \sum_{l=0}^L \mathbbm{1}(m_l \notin \mathcal S) }
            R^{\mathbf m}
            $ & $\breve{\mathbf  r}^{ (l, m_{l})} =
      \sum_{m_{l+1} \in \mathcal G}
    \alpha^{\mathbbm{1}(m_{l+1} \notin \mathcal{S})} 
    \mathbf T^{l, m_l, m_{l+1}}  \breve{   \mathbf  r}^{ (l+1, m_{l+1})}$ \\
\bottomrule
\end{tabular}
\label{table:GeneralLRPTable}
\end{table*}

\section{Forward-hook Trick}
\label{sec:ForwardHook}

We can implement our sGNN-LRP by slightly modifying the forward-hook trick \cite{schnake2020higher, samek2021explaining}, developed for the original GNN-LRP (see Appendix~\ref{FHDetail}).
We implement the forward combine step in Eq.\eqref{eq:GNN_agg_comb} as
\begin{equation}
    \begin{aligned}
        \mathbf P^{(l)} &\leftarrow \mathbf Z^{(l)} \mathbf W^{(l)\uparrow}, \\ 
        \mathbf Q^{(l)} &\leftarrow \mathbf P^{(l)} \odot [\sigma(\mathbf Z^{(l)}\mathbf W^{(l)}) \oslash \mathbf P^{(l)}]_{\texttt{cst.}}, \\ 
        \mathbf H^{(l)} &\leftarrow \mathbf Q^{(l)} \odot \mathbf M^{(l)} + [\mathbf Q^{(l)}]_{\texttt{cst.}} \odot(1-\mathbf M^{(l)}) ,
    \end{aligned}
    \label{eq:GCN_GNN_LRP_subgraph_rel_forward_hook_part1}
\end{equation}
where $\mathbf P^{(l)}, \mathbf M^{(l)} \in \mathbb R^{M \times N}$, and $\odot$ and $\oslash$ denote the entry-wise multiplication and division, respectively.
The operator $[\cdot]_{\texttt{cst.}}$ \emph{detaches} the quantity to which it applies from the gradient computation.  Then, the target quantity can be computed by the $\texttt{Autograd}$ function:
\begin{theorem}\label{thm:forward_hook}
Assume that we applied a complete forward prediction with the modified combine step \eqref{eq:GCN_GNN_LRP_subgraph_rel_forward_hook_part1} with the constant mask matrix $\mathbf M^{(l)} = \mathbf M^{\mathcal S}$ for all $l$, where $\mathbf M^{\mathcal S}$ masks the columns that correspond to the nodes outside the subgraph, i.e., the $m$-th columns for $m \in \mathcal S$ are all-one vectors, and the other columns are all-zero vectors.  Then, we get
$        R^{\mathcal  S} =\textstyle  \left<\texttt{Autograd}(y,\mathbf H^{(0)}), \mathbf M^{\mathcal S} \mathbf H^{(0)}\right>,
$
where $\left<\cdot, \cdot \right>$ denotes the Frobenius inner product.
\end{theorem}
The proof (given in Appendix  \ref{sec:ProofFH}) is similar to the one for the relevance of walk in \citet{schnake2020higher}.
This implementation 
is simpler, faster, and less memory intensive than the direct implementation of the sGNN-LRP rule \eqref{eq:GCN_GNN_LRP_rule_2}.

\section{Generalized Subgraph Attribution}
\label{sec:GeneralizedSubgraphAttribution}

In this section, we propose a novel definition of subgraph attribution by generalizing Eq.\eqref{eq:subgraph_rel_def_1},
and derive the corresponding LRP rule.  The proposed definition itself will be shown to be useful in Section~\ref{sec:Experiment}, and our derivation of a new propagation rule demonstrates the  utility of the message passing framework, introduced in Section~\ref{sec:LRPasMP}.

We consider the following two properties important to fulfill for subgraph attributions: 
A subgraph $\mathcal{S}$ is important if and only if
\begin{enumerate}
    \item the model makes almost the same predictions for the input graphs $\mathcal{G}$ and $\mathcal{S}$, and 
    \item the model predictions for its complement $\mathcal{G} \setminus \mathcal{S}$ and the full input graph $\mathcal{G}$ diverge drastically.
\end{enumerate}
However, 
the original definition \eqref{eq:subgraph_rel_def_1} of the subgraph attribution completely ignores the walks that step out of the subgraph even only once, and thus only considers the first property.

We propose a generalized version of subgraph attribution that trades-off both properties with a \emph{discounting} parameter $\alpha \in [0,1]$: 
\begin{equation}
    \begin{aligned}
        R_\alpha^{\mathcal S} = \textstyle \sum_{\mathbf m \in \mathcal G}
        g_\alpha^{\mathcal S}(\mathbf m) 
            R^{\mathbf m},
    \end{aligned}
    \label{eq:subgraph_rel_def_2}
\end{equation}
where 
\begin{equation}
    \begin{aligned}
        g_\alpha^{\mathcal S}(\mathbf m) = 
            \begin{cases}
                0 & \text{ if } m_l \notin \mathcal S, \ \forall \ l = 0, \ldots, L ,\\
       \alpha^{ \sum_{l=0}^L \mathbbm{1}(m_l \notin \mathcal S) } & \text{ otherwise}.\\
            \end{cases}
    \end{aligned}
    \label{eq:subgraph_rel_func_g}
\end{equation}
Here we used the indicator function $\mathbbm{1}(\cdot)$ equal to one if the event is true and zero otherwise.
The generalized subgraph attribution \eqref{eq:subgraph_rel_def_2} 
counts all walks that go through a node in $\mathcal{S}$ at least once with their discounted contributions according to how many times the walk steps outside $\mathcal{S}$.  
For $\alpha=0$, it reduces to the original definition, i.e., $R^{\mathcal{S}} = R_0^{\mathcal{S}} $.

\begin{table*}[t]
    \caption{Computation time (in msec) comparison on 5 datasets. '---' means `failed'.%
    The subgraph size is $|\mathcal{S}| = 5$.}    \label{tab:efficiency_table}
\vspace{-4mm}
    \begin{center}
        \begin{small}
            \begin{sc}
                \begin{tabular}{llccccccc}
                    \toprule
                    \multicolumn{2}{l}{Dataset} & \multicolumn{3}{c}{BA-2motif} & MUTAG & Mutagenicity & REDDIT-B & Graph-SST2 \\ \midrule
\multicolumn{2}{l}{
Model-$L$ (depth)} 
& GIN-3 & GIN-5 & GIN-7 & GIN-3 & GIN-3 & GIN-5 & GCN-3 \\ \midrule
Naive GNN-LRP &  & 224.22 & $6.07 \!\times\! 10^{3}$ & $1.42 \!\times\! 10^{5}$ & $4.23 \!\times\! 10^{3}$ & $4.28 \!\times\! 10^{3}$ & --- & $3.16 \!\times\! 10^{5}$ \\ 
sGNN-LRP (Ours) &  & {\bf 4.22} & {\bf 6.44} & {\bf 9.81} & {\bf 28.90} & {\bf 26.68} & {\bf 195.43} &{\bf  29.94} \\

                    \bottomrule
                \end{tabular}
            \end{sc}
        \end{small}
    \end{center}
\end{table*}

We can efficiently compute the generalized subgraph attribution \eqref{eq:subgraph_rel_def_2} by decomposing it as 
\begin{equation}
    \begin{aligned}
        R_\alpha^{\mathcal S} = \widetilde{R}_\alpha^{\mathcal S}
        - \alpha^{L+1}  R_0^{\mathcal{G} \setminus\mathcal{S} },
    \end{aligned}
    \label{eq:subgraph_rel_decomposition}
\end{equation}
where 
\begin{equation}
    \begin{aligned}
        \widetilde{R}_\alpha^{\mathcal S} = \textstyle \sum_{\mathbf m \in \mathcal G}
       \alpha^{ \sum_{l=0}^L \mathbbm{1}(m_l \notin \mathcal S) }
            R^{\mathbf m},
    \end{aligned}
    \label{eq:subgraph_rel_def_3}
\end{equation}
and applying a message passing algorithm for Eq.\eqref{eq:subgraph_rel_def_3}.
Note that the second term in Eq.\eqref{eq:subgraph_rel_decomposition} is the original ($\alpha = 0$) subgraph attribution to the complementary set of $\mathcal{S}$ weighted by $\alpha^{L+1}$, which can be efficiently computed by sGNN-LRP, described in Section~\ref{sec:MPAlgorithmProposal}.
For the first term (or Eq.\eqref{eq:subgraph_rel_def_3}), our message passing framework, introduced in Section~\ref{sec:LRPasMP}, gives the following theorem (the proof is given in Appendix~\ref{sec:ProofSP}):
\begin{theorem}
\label{thm:sumprod_general}
Assume that we apply the LRP rule
\begin{align}
    \begin{aligned}
        &
        \breve{\mathbf  r}^{(l, m_l)}        = \\
        & \quad \textstyle \sum_{m_{l+1} \in \mathcal G}
    \alpha^{\mathbbm{1}(m_{l+1} \notin \mathcal{S})} \,
    \mathbf T^{l, m_l, m_{l+1}} \, 
       \breve{\mathbf  r}^{(l+1, m_{l+1})} 
    \end{aligned}
    \label{eq:GCN_GNN_LRP_rule_general}
\end{align}
for $l = L-1, \ldots, 0$ with the initial message $\breve{\mathbf{r}}^{(L, m_L)} = \mathbf{r}^{L, m_L}$.  Then, $  \breve{\mathbf  r}^{(l, m_l)} = \widetilde{\mathbf r}^{\, l, m_l, m_{l+1: L } \subseteq \mathcal G} \; \forall l \in \{0, \ldots, L\}  $, 
where
$\widetilde{\mathbf r}^{\, l, m_l, m_{l+1: L } \subseteq \mathcal G}  =  \alpha^{ \sum_{l'=l+1}^L \mathbbm{1}(m_{l'} \notin \mathcal S) }
{\mathbf r}^{\, l, m_l, m_{l+1: L } \subseteq \mathcal G} $,
and 
\[
\widetilde{R}_\alpha^{\mathcal S} =\textstyle  \sum_{m_{0}\in \mathcal G}  \alpha^{\mathbbm{1}(m_{0} \notin \mathcal{S})} \sum_{n=1}^{N^{(0)}}  \widetilde{r}_n^{\, 0, m_0, m_{1: L } \subseteq \mathcal G}.
\]
\end{theorem}
The forward-hook trick is also applicable
(the proof is given in Appendix~\ref{sec:ProofFH}):
\begin{theorem}\label{thm:forward_hook_general}
Assume that we applied a complete forward prediction with the modified combine step \eqref{eq:GCN_GNN_LRP_subgraph_rel_forward_hook_part1} with the constant mask matrix $\mathbf M^{(l)} = \mathbf M_\alpha^{\mathcal S}$ for all $l$, where $\mathbf M_\alpha^{\mathcal S}$ \emph{softly} masks the nodes outside the subgraph, i.e.,  the columns corresponding to the nodes in the subgraph $\mathcal S$ are all-one vectors, and all the other entries are equal to $\alpha$.  Then, we get
 $       \widetilde{R}_\alpha^{\mathcal  S} = \textstyle \left<\texttt{Autograd}(y,\mathbf H^{(0)}), \mathbf M_\alpha^{\mathcal S} \mathbf H^{(0)}\right>
 $.
\end{theorem}
We will show the usefulness of the generalized subgraph attribution in Section~\ref{sec:Experiment}.

\section{Experiment}
\label{sec:Experiment}

In this section, we conduct two experiments demonstrating (1) the massive gain in computation time by our efficient sGNN-LRP,
and (2) the usefulness of the generalized subgraph attribution in relevant node-ordering tasks.
For GNN models, we used GIN and GCN with different depths $L$.
We used the following five popular datasets: 
\textbf{BA-2motif} \cite{luo2020parameterized},
\textbf{MUTAG} \cite{debnath1991structure},
\textbf{Mutagenicity} \cite{doi:10.1021/jm040835a},
\textbf{REDDIT-BINARY} \cite{DBLP:conf/kdd/YanardagV15},
and \textbf{Graph-SST2} \cite{yuan2020explainability}.
Detailed experimental setting is given in Appendix~\ref{sec:ModelDetails},
and our implementation is available at our GitHub repository.%
\footnote{
\url{https://github.com/xiong-ping/sgnn_lrp_via_mp}.
}

\subsection{Computational Efficiency Evaluation}

Here we show computational advantages of sGNN-LRP over the  Naive GNN-LRP \citep{schnake2020higher} as a baseline on different scales of models and subgraph sizes.
Experiments were performed on a Xeon E5-2620 CPU with 8GB memory. 

\begin{figure}[t]
\begin{center}
  \subfigure[Network depth dependence]{\includegraphics[scale=0.43]{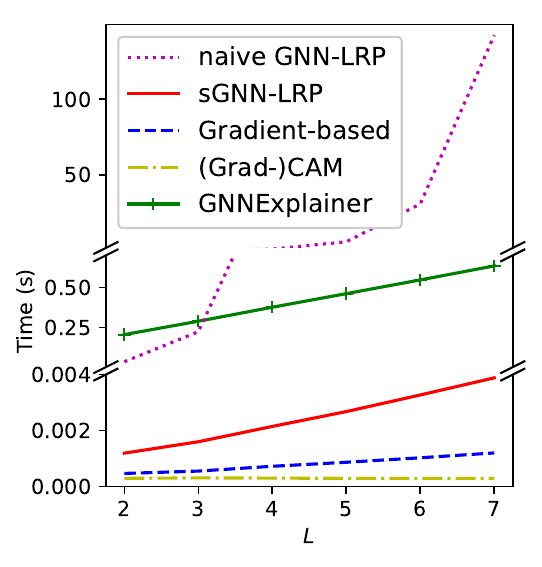}\label{fig:runtime_baselines_L}} \quad
  \subfigure[Subgraph size dependence]{\includegraphics[scale=0.43]{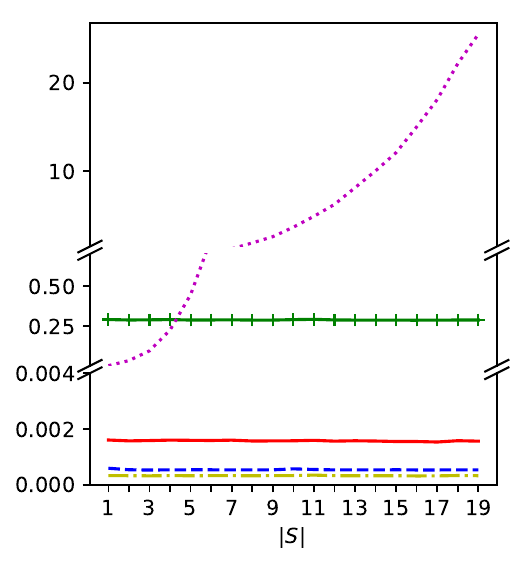}\label{fig:runtime_baselines_S}}
\vspace{-3mm}
  \caption{Computation time on BA-2motif dataset. Note the different vertical scales in the top, middle, and bottom parts.
  (a) GIN-$L$ for $L = 1, \ldots, 6$ with  $|\mathcal{S}| = 5$.   (b) GIN-$3$ with  $|\mathcal{S}| = 1, \ldots, 19$. 
  }
  \vspace{-3mm}
\label{fig:runtime_baselines}
\end{center}
\end{figure}

Figure~\ref{fig:runtime_baselines}
shows the computation time for subgraph attribution on BA-2motif, as functions of (a) the network depth $L$ and (b) the subgraph size $|\mathcal{S}|$, respectively. 
We clearly observe the (a) exponential and the (b) cubic ($L=3$) complexity, $O(|\mathcal S|^L)$, of Naive GNN-LRP.
Our proposed sGNN-LRP with its complexity $O(L |\mathcal{S}|^2)$ is drastically faster. 
Table \ref{tab:efficiency_table} summarizes the computation time for various GNNs and datasets.
The reported computation time is the average over three trials for randomly chosen 50 samples in each dataset.  We report `fail' if out-of-memory error occurs or the computation does not finish within 360 sec. Again we observe from the table a significant computational gain by  sGNN-LRP.

We also compared computation time with other baseline methods in Figure~\ref{fig:runtime_baselines},
and observed that our sGNN-LRP is significantly faster than GNNExplainer, and even comparable with very simple baselines, Gradient-based heatmap and (Grad-)CAM.
Notably the complexity bounds $O(L |\mathcal{S}|^2)$ of sGNN-LRP is the same as those for a single forward/backward pass of GNNs, and therefore, sGNN-LRP can be applied to deeper GNNs for larger graphs at a similar computational cost to prediction.

\subsection{Node Ordering Performance by Generalized Subgraph Attribution}

Here, we demonstrate the high usefulness of our generalized definition of subgraph attribution.  Specifically, we show that the optimal discounting parameter, $\alpha$ in Eq.\eqref{eq:subgraph_rel_func_g}, is not always zero,
and depends on the evaluation task.
We first evaluate the node ordering performance on the BA-2motif dataset, for which the ground truth is available.
Then, we evaluate the performance in the model activation and the model pruning tasks \citep{schnake2020higher} on other datasets.

\subsubsection{Node Ordering}
\label{sec:NodeOrdering}

We use the generalized subgraph attribution for providing node ordering in two modes, \emph{activation} and \emph{pruning}.

\textbf{Activation mode:}
In this mode, we obtain the node ordering $\left[m^{(1)},...,m^{(M)}\right]$ such that
$\sum_{i=1}^{M} R_{\alpha}^{\{m^{(1)},...,m^{(i)}\}}$ is maximized.  
Since this maximization is infeasible, we perform the following greedy search for $i = 1, \cdots, M$:
$        m^{(i)} = \underset{m \in \mathcal G \backslash \mathcal \{m^{(1)},...,m^{(i-1)}\}} {\mathrm{argmax}}  R_\alpha^{ \{m^{(1)},...,m^{(i-1)},  m\}  } 
  $.

\textbf{Pruning mode:}
In this mode, we obtain the node ordering $\left[m^{(1)},...,m^{(M)}\right]$ such that
$\sum_{i=1}^{M} |R_\alpha^{\mathcal G \backslash \{m^{(1)},...,m^{(i)}\}}-R^{\mathcal G}|$ is minimized.  
This minimization is again infeasible, and therefore, we perform the following greedy search for $i = 1, \cdots, M$:
$        m^{(i)} = \underset{m \in \mathcal G \backslash \mathcal \{m^{(1)},...,m^{(i-1)}\}} {\mathrm{argmin}} | R_\alpha^{\mathcal{G} \setminus  \{m^{(1)},...,m^{(i-1)}, m\}  } 
        - R^{\mathcal{G}}
        |$.

Since the activation mode focuses on the high relevance subgraphs themselves and the pruning mode focuses on their complement, 
we expect that they respectively tend to satisfy the first
and the second properties that good  subgraph attribution should fulfill (see Section~\ref{sec:GeneralizedSubgraphAttribution}).
We investigate how the discounting parameter $\alpha$ of the generalized subgraph attribution affects the activation and the pruning performance.

\subsubsection{BA-2motif Benchmark Experiment}

\begin{figure}[t]
\begin{center}
\centerline{\includegraphics[scale=0.5]{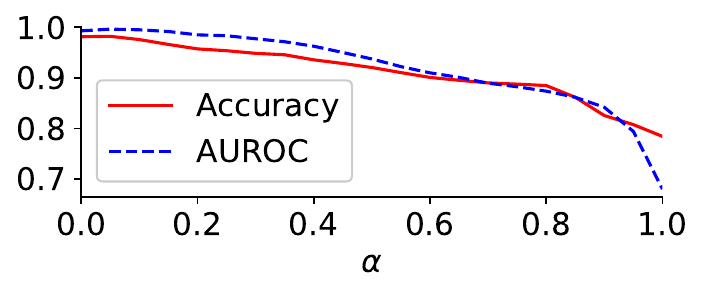}}
\caption{Accuracy and AUROC of node ordering task on the BA-2motif dataset with different discounting parameter $\alpha$. The best performance is achieved around $\alpha \sim 0$.}
\label{fig:ba2motif_acc_auc}
\end{center}
\end{figure}

Here, we use the BA-2motif dataset to
evaluate the node ordering performance of subgraph attribution.
Since this dataset is synthetic and the ground truth \emph{motif} is available, 
we can simply compare the node ordering obtained by the subgraph attribution
with the ground truth.
All sample graphs have 25 nodes, of which 5 nodes are specified as the motif. 
Fig. \ref{fig:ba2motif_acc_auc} shows the accuracy---the proportion that the subgraph attribution in the activation mode gives the ordering such that the top 5 nodes match the ground truth motif---with its dependence on $\alpha$.
The figure also shows the area under the receiver operating characteristic curve (AUROC),
where the threshold of motif detection is scanned.
We found that the best performance is achieved around $\alpha \sim 0$, and therefore, the original subgraph attribution is almost optimal.
This is not surprising  
because, for this dataset, the nodes outside the motifs are completely random, and therefore, considering the outside nodes gives no useful information.

\subsubsection{Model Activation and Pruning Experiments}
\label{sec:APExperiment}

\begin{figure*}[t]
\begin{center}
    \includegraphics[width=\linewidth]{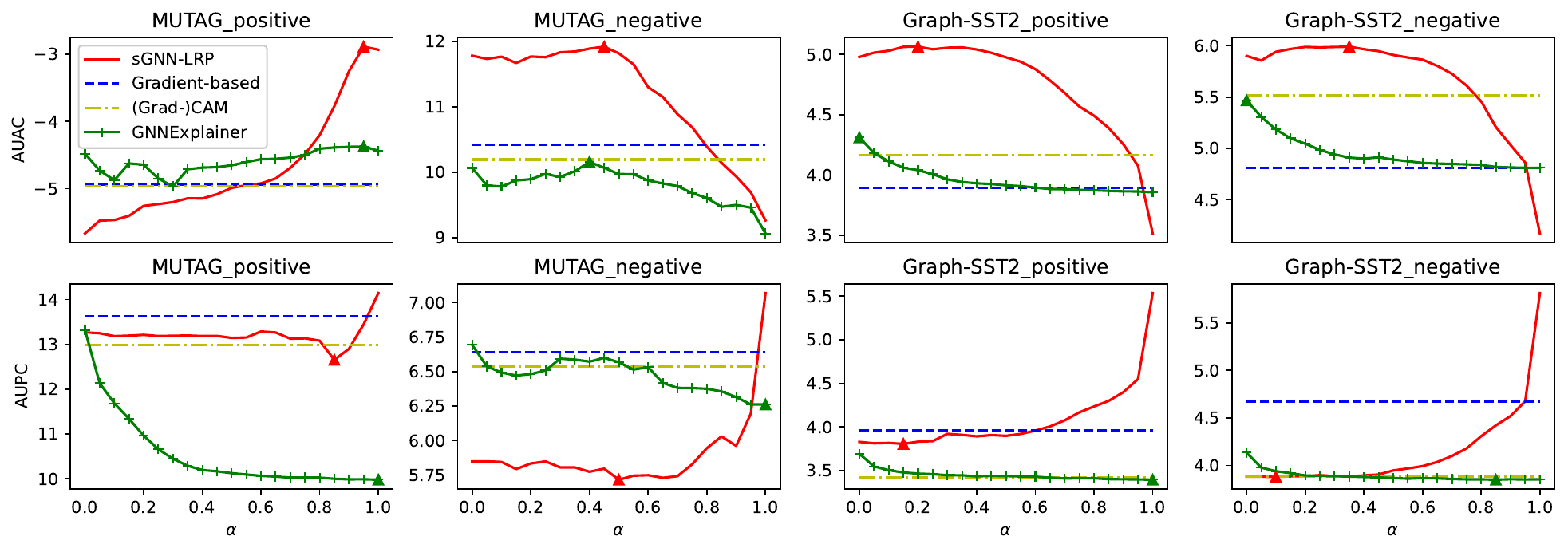}
\caption{
AUAC (top row, higher is better) and AUPC (bottom row, lower is better) on MUTAG and Graph-SST2 datasets of our sGNN-LRP and baseline methods, GNNExplainer, Gradient-based heatmap, and (Grad-)CAM.
The triangles mark the best performance points for sGNN-LRP and GNNExplainer.
}

\label{fig:auac_aupc_result}
\end{center}
\end{figure*}

Next we evaluate the subgraph attribution performance in model activation and model pruning tasks on MUTAG and Graph-SST2 datasets, for which the ground truth explanation is not available.
These tasks measure the correlation between the relevance attribution and the model output in two ways.

Let $f(\cdot)$ be the GNN model output for the correct label.
The goal of the model activation task is to find the node ordering 
$\left[m^{(1)},...,m^{(M)}\right]$ such that
the area under the activation curve, 
$\textrm{AUAC} =
\textstyle 
\frac{1}{M}\sum_{i=1}^{M} f(\{m^{(1)},...,m^{(i)}\})
$,
is maximized. This task evaluates how many subgraphs formed by the top predicted nodes recover the correct prediction, and therefore, measures how well the attribution fulfills Property 1 in Section~\ref{sec:GeneralizedSubgraphAttribution}.  
On the other hand, the goal of the pruning task is to find the node ordering 
$\left[m^{(1)},...,m^{(M)}\right]$ such that
the area under the pruning curve,
$\textrm{AUPC} =
\textstyle 
\frac{1}{M}\sum_{i=1}^{M} | f(\mathcal{G} \setminus \{m^{(1)},...,m^{(i)}\}) - f(\mathcal{G}) |
$,
is minimized.
This task evaluates how much the complement of subgraphs formed by the top predicted nodes  retains the predictive performance,
and therefore, can be a performance measure on Property 2 in Section~\ref{sec:GeneralizedSubgraphAttribution}.  
Further details are given in Appendix \ref{sec:detail_acti_prun}.
We should naturally use the activation and the pruning modes,
respectively, for node ordering
in the model activation and the model pruning tasks (see Section~\ref{sec:NodeOrdering}).

Figure~\ref{fig:auac_aupc_result} shows AUAC and AUPC with their dependence on the discounting parameter $\alpha$.
Because AUAC, as well as AUPC, differs largely between positive samples and negative samples, we plotted them separately. We speculate that this is due to the different predictive capability of the model on positive and negative samples. The (red) curves for sGNN-LRP imply that the original definition of the subgraph attribution, i.e.,  $\alpha = 0$, is not always optimal, and tuning $\alpha$ can improve the node ordering performance.  This applies not only in the pruning task but also in the activation task.  We will further investigate the performance dependence on $\alpha$, and develop tuning procedures in our future work.

To show the superiority of the attribution via sGNN-LRP, we also compared the node ordering performance  with three baselines, GNNExplainer, Gradient-based heatmap, and (Grad-)CAM, and plotted the results in Figure~\ref{fig:auac_aupc_result}.
The subgraph relevance by GNNExplainer for $\alpha=0$ is given by the sum of the relevances over the edges between the nodes both inside the subgraph.  For $\alpha>0$, the sum over the edges connecting inside and outside nodes is also added with the discounting factor $\alpha$ (see Appendix~\ref{sec:gnn_alpha} for more detail).
Assuming that the parameter $\alpha$ is optimized for each method, sGNN-LRP outperforms the baselines in most of the cases (6 out of 8).

\section{Conclusion}

Layer-wise relevance propagation for
Graph neural networks
(GNN-LRP) is a higher-order explainability method for GNNs, which provides attributes of the GNN models at the level of walks.  Specifically, it  supports subgraph-level attribution by summing over the walks inside a given subgraph, which however suffers from {\em exponential} complexity and thus has computational limits in application. In this paper, we
have overcome this issue by proposing a {\em polynomial-time} algorithm (sGNN-LRP) that directly computes the subgraph GNN-LRP attribution.
Notably, 
our development of sGNN-LRP has been conducted by a novel procedure: 
unlike previous work, we developed sGNN-LRP by first defining the target quantity to be computed and then deriving a propagation rule as a message passing algorithm.
This novel procedure is general, rediscovering many existing LRP rules, and thus expected to facilitate future development of new LRP methods.
We have demonstrated the utility of this procedure
by deriving another LRP rule for computing a  generalized definition of the subgraph relevance
 that takes into account the partly-outside walks.
Experimental results showed that our proposed sGNN-LRP is significantly faster than the naive application of GNN-LRP,
 and that the generalized subgraph relevance definition can more robustly attribute GNNs at the subgraph level.
 
 Future research will address the broad application of the novel algorithms, as now novel `deeper' insights (manifested in longer walks or deeper GNNs) have become possible for learning problems that possess a significant amount of higher-order and long range nonlinear interactions, such as in the sciences, e.g.,~for neuroimaging (see, e.g.,~\citet{rubinov2010complex,shine2019human}) or quantum chemistry (see, e.g.,~\citet{gilmer2017neural,schutt2018schnet,unke2021spookynet}). Beyond applications in the sciences, we consider the novel efficient GNN-LRP algorithms as promising for NLP applications where assessing deeper higher-order interactions may be helpful for assessing trustworthiness, fairness, and unbiasedness of SOTA systems.

\section*{Acknowledgments}
We would like to thank Michael Gastegger for helpful discussion and his significant support in preparing the manuscript after the abstract submission deadline. 
This work was supported by the German Ministry for Education and Research (BMBF) as BIFOLD - Berlin Institute for the Foundations of Learning and Data under grants 01IS18025A and  01IS18037A.

\bibliography{main}

@article{wu2020comprehensive,
  author    = {Zonghan Wu and
               Shirui Pan and
               Fengwen Chen and
               Guodong Long and
               Chengqi Zhang and
               Philip S. Yu},
  title     = {A Comprehensive Survey on Graph Neural Networks},
  journal   = {{IEEE} Trans. Neural Networks Learn. Syst.},
  volume    = {32},
  number    = {1},
  pages     = {4--24},
  year      = {2021} 
}

@article{bach2015pixel,
  title={On pixel-wise explanations for non-linear classifier decisions by layer-wise relevance propagation},
  author={Bach, Sebastian and Binder, Alexander and Montavon, Gr{\'e}goire and Klauschen, Frederick and M{\"u}ller, Klaus-Robert and Samek, Wojciech},
  journal={PloS one},
  volume={10},
  number={7},
  pages={e0130140},
  year={2015},
  publisher={Public Library of Science San Francisco, CA USA}
}

@article{rubinov2010complex,
  title={Complex network measures of brain connectivity: uses and interpretations},
  author={Rubinov, Mikail and Sporns, Olaf},
  journal={Neuroimage},
  volume={52},
  number={3},
  pages={1059--1069},
  year={2010},
  publisher={Elsevier}
}

@article{unke2021spookynet,
    Author = {Unke, Oliver T. and Chmiela, Stefan and Gastegger, Michael and Schütt, Kristof T. and Sauceda, Huziel E. and Müller, Klaus-Robert},
	doi = {10.1038/s41467-021-27504-0},
	isbn = {2041-1723},
	journal = {Nature Communications},
	number = {1},
	pages = {7273},
	title = {SpookyNet: Learning force fields with electronic degrees of freedom and nonlocal effects},
	ty = {JOUR},
	url = {https://doi.org/10.1038/s41467-021-27504-0},
	volume = {12},
	year = {2021}
}

@article{shine2019human,
  title={Human cognition involves the dynamic integration of neural activity and neuromodulatory systems},
  author={Shine, James M and Breakspear, Michael and Bell, Peter T and Martens, Kaylena A Ehgoetz and Shine, Richard and Koyejo, Oluwasanmi and Sporns, Olaf and Poldrack, Russell A},
  journal={Nature neuroscience},
  volume={22},
  number={2},
  pages={289--296},
  year={2019},
  publisher={Nature Publishing Group}
}

@article{montavon2018methods,
  title={Methods for interpreting and understanding deep neural networks},
  author={Montavon, Gr{\'e}goire and Samek, Wojciech and M{\"u}ller, Klaus-Robert},
  journal={Digital Signal Processing},
  volume={73},
  pages={1--15},
  year={2018},
  publisher={Elsevier}
}

@article{samek2021explaining,
  author    = {Wojciech Samek and
               Gr{\'{e}}goire Montavon and
               Sebastian Lapuschkin and
               Christopher J. Anders and
               Klaus-Robert M{\"{u}}ller},
  title     = {Explaining Deep Neural Networks and Beyond: {A} Review of Methods
               and Applications},
  journal   = {Proc. {IEEE}},
  volume    = {109},
  number    = {3},
  pages     = {247--278},
  year      = {2021} 
}

@inbook{bishop2006patternchp8, 
    author = {Bishop, Christopher M.}, 
    title = {Pattern Recognition and Machine Learning (Information Science and Statistics)}, 
    year = {2006}, 
    isbn = {0387310738}, 
    publisher = {Springer-Verlag}, 
    address = {Berlin, Heidelberg},
    pages = {402--411}
}

@article{yuan2020explainability,
  author    = {Hao Yuan and
               Haiyang Yu and
               Shurui Gui and
               Shuiwang Ji},
  title     = {Explainability in Graph Neural Networks: {A} Taxonomic Survey},
  journal   = {CoRR},
  volume    = {abs/2012.15445},
  year      = {2020},
  url       = {https://arxiv.org/abs/2012.15445},
  eprinttype = {arXiv},
  eprint    = {2012.15445}
}

@inproceedings{ying2019gnnexplainer,
  author    = {Zhitao Ying and
               Dylan Bourgeois and
               Jiaxuan You and
               Marinka Zitnik and
               Jure Leskovec},
  editor    = {Hanna M. Wallach and
               Hugo Larochelle and
               Alina Beygelzimer and
               Florence d'Alch{\'{e}}{-}Buc and
               Emily B. Fox and
               Roman Garnett},
  title     = {GNNExplainer: Generating Explanations for Graph Neural Networks},
  booktitle = {Advances in Neural Information Processing Systems 32: Annual Conference
               on Neural Information Processing Systems 2019, NeurIPS 2019, December
               8-14, 2019, Vancouver, BC, Canada},
  pages     = {9240--9251},
  year      = {2019} 
}

@inproceedings{luo2020parameterized,
  author    = {Dongsheng Luo and
               Wei Cheng and
               Dongkuan Xu and
               Wenchao Yu and
               Bo Zong and
               Haifeng Chen and
               Xiang Zhang},
  editor    = {Hugo Larochelle and
               Marc'Aurelio Ranzato and
               Raia Hadsell and
               Maria{-}Florina Balcan and
               Hsuan{-}Tien Lin},
  title     = {Parameterized Explainer for Graph Neural Network},
  booktitle = {Advances in Neural Information Processing Systems 33: Annual Conference
               on Neural Information Processing Systems 2020, NeurIPS 2020, December
               6-12, 2020, virtual},
  year      = {2020} 
}

@inproceedings{yuan2021explainability,
  author    = {Hao Yuan and
               Haiyang Yu and
               Jie Wang and
               Kang Li and
               Shuiwang Ji},
  editor    = {Marina Meila and
               Tong Zhang},
  title     = {On Explainability of Graph Neural Networks via Subgraph Explorations},
  booktitle = {Proceedings of the 38th International Conference on Machine Learning,
               {ICML} 2021, 18-24 July 2021, Virtual Event},
  series    = {Proceedings of Machine Learning Research},
  volume    = {139},
  pages     = {12241--12252},
  publisher = {{PMLR}},
  year      = {2021} 
}

@inproceedings{vu2020pgm,
  author    = {Minh N. Vu and
               My T. Thai},
  editor    = {Hugo Larochelle and
               Marc'Aurelio Ranzato and
               Raia Hadsell and
               Maria{-}Florina Balcan and
               Hsuan{-}Tien Lin},
  title     = {{PGM}-{E}xplainer: Probabilistic Graphical Model Explanations for Graph
               Neural Networks},
  booktitle = {Advances in Neural Information Processing Systems 33: Annual Conference
               on Neural Information Processing Systems 2020, NeurIPS 2020, December
               6-12, 2020, virtual},
  year      = {2020} 
}

@inproceedings{xu2018powerful,
  author    = {Keyulu Xu and
               Weihua Hu and
               Jure Leskovec and
               Stefanie Jegelka},
  title     = {How Powerful are Graph Neural Networks?},
  booktitle = {7th International Conference on Learning Representations, {ICLR} 2019,
               New Orleans, LA, USA, May 6-9, 2019},
  year      = {2019} 
}

@article{schnake2020higher,
  author={Schnake, Thomas and Eberle, Oliver and Lederer, Jonas and Nakajima, Shinichi and Sch{\"u}tt, Kristof T and M{\"u}ller, Klaus-Robert and Montavon, Gregoire},
  journal={IEEE Transactions on Pattern Analysis and Machine Intelligence}, 
  title={Higher-Order Explanations of Graph Neural Networks via Relevant Walks}, 
  year={2021},
  volume={},
  number={},
  pages={1-1} }

@inproceedings{chen2018fastgcn,
  author    = {Jie Chen and
               Tengfei Ma and
               Cao Xiao},
  title     = {Fast{GCN}: Fast Learning with Graph Convolutional Networks via Importance
               Sampling},
  booktitle = {6th International Conference on Learning Representations, {ICLR} 2018,
               Vancouver, BC, Canada, April 30 - May 3, 2018, Conference Track Proceedings},
  year      = {2018} 
}

@INPROCEEDINGS{domingue2019evolution,
  author={Domingue, Miguel and Dhamdhere, Rohan and Harish Kanamarlapudi, Naga Durga and Raghupathi, Sunand and Ptucha, Raymond},
  booktitle={2019 IEEE Western New York Image and Signal Processing Workshop (WNYISPW)}, 
  title={Evolution of Graph Classifiers}, 
  year={2019},
  volume={},
  number={},
  pages={1-5} }

@inproceedings{hamilton2017inductive,
  author    = {William L. Hamilton and
               Zhitao Ying and
               Jure Leskovec},
  editor    = {Isabelle Guyon and
               Ulrike von Luxburg and
               Samy Bengio and
               Hanna M. Wallach and
               Rob Fergus and
               S. V. N. Vishwanathan and
               Roman Garnett},
  title     = {Inductive Representation Learning on Large Graphs},
  booktitle = {Advances in Neural Information Processing Systems 30: Annual Conference
               on Neural Information Processing Systems 2017, December 4-9, 2017,
               Long Beach, CA, {USA}},
  pages     = {1024--1034},
  year      = {2017} 
}

@article{schutt2018schnet,
  title={Schnet--a deep learning architecture for molecules and materials},
  author={Sch{\"u}tt, Kristof T and Sauceda, Huziel E and Kindermans, P-J and Tkatchenko, Alexandre and M{\"u}ller, K-R},
  journal={The Journal of Chemical Physics},
  volume={148},
  number={24},
  pages={241722},
  year={2018},
  publisher={AIP Publishing LLC}
}

@article{DBLP:journals/tnn/ScarselliGTHM09,
  author    = {Franco Scarselli and
               Marco Gori and
               Ah Chung Tsoi and
               Markus Hagenbuchner and
               Gabriele Monfardini},
  title     = {The Graph Neural Network Model},
  journal   = {{IEEE} Trans. Neural Networks},
  volume    = {20},
  number    = {1},
  pages     = {61--80},
  year      = {2009} 
}

@inproceedings{DBLP:conf/iclr/KipfW17,
  author    = {Thomas N. Kipf and
               Max Welling},
  title     = {Semi-Supervised Classification with Graph Convolutional Networks},
  booktitle = {5th International Conference on Learning Representations, {ICLR} 2017,
               Toulon, France, April 24-26, 2017, Conference Track Proceedings},
  year      = {2017} 
}

@inproceedings{DBLP:conf/kdd/YuanTHJ20,
  author    = {Hao Yuan and
               Jiliang Tang and
               Xia Hu and
               Shuiwang Ji},
  editor    = {Rajesh Gupta and
               Yan Liu and
               Jiliang Tang and
               B. Aditya Prakash},
  title     = {{XGNN:} Towards Model-Level Explanations of Graph Neural Networks},
  booktitle = {{KDD} '20: The 26th {ACM} {SIGKDD} Conference on Knowledge Discovery
               and Data Mining, Virtual Event, CA, USA, August 23-27, 2020},
  pages     = {430--438},
  publisher = {{ACM}},
  year      = {2020} 
}

@inproceedings{DBLP:conf/cvpr/PopeKRMH19,
  author    = {Phillip E. Pope and
               Soheil Kolouri and
               Mohammad Rostami and
               Charles E. Martin and
               Heiko Hoffmann},
  title     = {Explainability Methods for Graph Convolutional Neural Networks},
  booktitle = {{IEEE} Conference on Computer Vision and Pattern Recognition, {CVPR}
               2019, Long Beach, CA, USA, June 16-20, 2019},
  pages     = {10772--10781},
  publisher = {Computer Vision Foundation / {IEEE}},
  year      = {2019} 
}

@inproceedings{DBLP:conf/aaai/Pearl82,
  author    = {Judea Pearl},
  editor    = {David L. Waltz},
  title     = {Reverend {B}ayes on Inference Engines: {A} Distributed Hierarchical
               Approach},
  booktitle = {Proceedings of the National Conference on Artificial Intelligence,
               Pittsburgh, PA, USA, August 18-20, 1982},
  pages     = {133--136},
  publisher = {{AAAI} Press},
  year      = {1982}}

@article{debnath1991structure,
author = {Debnath, Asim Kumar and Lopez de Compadre, Rosa L. and Debnath, Gargi and Shusterman, Alan J. and Hansch, Corwin},
title = {Structure-activity relationship of mutagenic aromatic and heteroaromatic nitro compounds. Correlation with molecular orbital energies and hydrophobicity},
journal = {Journal of Medicinal Chemistry},
volume = {34},
number = {2},
pages = {786-797},
year = {1991},
doi = {10.1021/jm00106a046}
}

@article{doi:10.1021/jm040835a,
author = {Kazius, Jeroen and McGuire, Ross and Bursi, Roberta},
title = {Derivation and Validation of Toxicophores for Mutagenicity Prediction},
journal = {Journal of Medicinal Chemistry},
volume = {48},
number = {1},
pages = {312-320},
year = {2005},
doi = {10.1021/jm040835a},
    note ={PMID: 15634026}
}

@inproceedings{DBLP:conf/kdd/YanardagV15,
  author    = {Pinar Yanardag and
               S. V. N. Vishwanathan},
  editor    = {Longbing Cao and
               Chengqi Zhang and
               Thorsten Joachims and
               Geoffrey I. Webb and
               Dragos D. Margineantu and
               Graham Williams},
  title     = {Deep Graph Kernels},
  booktitle = {Proceedings of the 21th {ACM} {SIGKDD} International Conference on
               Knowledge Discovery and Data Mining, Sydney, NSW, Australia, August
               10-13, 2015},
  pages     = {1365--1374},
  publisher = {{ACM}},
  year      = {2015},
  doi       = {10.1145/2783258.2783417}
}

@article{hu2020ogb,
  title={Open Graph Benchmark: Datasets for Machine Learning on Graphs},
  author={Hu, Weihua and Fey, Matthias and Zitnik, Marinka and Dong, Yuxiao and Ren, Hongyu and Liu, Bowen and Catasta, Michele and Leskovec, Jure},
  journal={arXiv preprint arXiv:2005.00687},
  year={2020}
}

@inproceedings{lund2017unified,
 author = {Lundberg, Scott M and Lee, Su-In},
 booktitle = {Advances in Neural Information Processing Systems},
 editor = {I. Guyon and U. V. Luxburg and S. Bengio and H. Wallach and R. Fergus and S. Vishwanathan and R. Garnett},
 publisher = {Curran Associates, Inc.},
 title = {A Unified Approach to Interpreting Model Predictions},
 volume = {30},
 pages = {4768–4777},
 year = {2017}
}

@inproceedings{gilmer2017neural,
author = {Gilmer, Justin and Schoenholz, Samuel S. and Riley, Patrick F. and Vinyals, Oriol and Dahl, George E.},
title = {Neural Message Passing for Quantum Chemistry},
year = {2017},
publisher = {JMLR.org},
booktitle = {Proceedings of the 34th International Conference on Machine Learning - Volume 70},
pages = {1263–1272},
numpages = {10},
location = {Sydney, NSW, Australia},
series = {ICML'17}
}

@article{Jeroen205derivmutag,
author = {Kazius, Jeroen and McGuire, Ross and Bursi, Roberta},
title = {Derivation and Validation of Toxicophores for Mutagenicity Prediction},
journal = {Journal of Medicinal Chemistry},
volume = {48},
number = {1},
pages = {312-320},
year = {2005}

}

@inproceedings{DBLP:conf/icdm/GaoSBYH021,
  author    = {Yuyang Gao and
               Tong Sun and
               Rishab Bhatt and
               Dazhou Yu and
               Sungsoo Hong and
               Liang Zhao},
  editor    = {James Bailey and
               Pauli Miettinen and
               Yun Sing Koh and
               Dacheng Tao and
               Xindong Wu},
  title     = {{GNES:} Learning to Explain Graph Neural Networks},
  booktitle = {{IEEE} International Conference on Data Mining, {ICDM} 2021, Auckland,
               New Zealand, December 7-10, 2021},
  pages     = {131--140},
  publisher = {{IEEE}},
  year      = {2021}
}
\bibliographystyle{icml2022}

\newpage
\appendix
\onecolumn

\section{Details of Figure~\ref{fig:comp_holo_mol}}\label{app:fig1}
In Figure \ref{fig:comp_holo_mol} we consider the GIN model trained on the Mutagenicity dataset \cite{doi:10.1021/jm040835a}. We use the same model architecture and training procedure as described in Appendix \ref{app:mutagenicity}. The visualized molecule is one sample of the Mutagenicity dataset which is classified as mutagenic. The bars in each graph reflect the interaction relevance of the subgraph consisting of all atoms with the described atomic number at the $x$-axis. In order to reflect the relevance of the interactions of a set $\mathcal{S}$, we use the subgraph relevance definition in equation \eqref{eq:subgraph_rel_def_1}  and subtract the relevance scores of all features that are not exclusively composed of all atoms types in $S$. For example, in the case where we measure the interaction between the atoms C, N and O we define the interaction relevance $R_{\leftrightarrow}^{CNO}$ to be
\begin{align*}
R_{\leftrightarrow}^{CNO} = R^{CNO} - R^{CN} - R^{CO} - R^{NO} + R^C +R^N +R^O.
\end{align*}
  It is important to see that we add the relevance scores with only one atom (such as $R^C$), because by definition of the subraph relevance in \eqref{eq:subgraph_rel_def_1} they occur in the scores which consists of two atoms (as in $R^{CO}$ and $R^{CN}$), which we already subtract twice. This definition ensures the conservation of the interaction relevance, i.e. it ensures $R^{CNO} = \sum_{\mathcal{S} \, \subset \, CNO} R^\mathcal{S}_{\leftrightarrow}$.

\section{Proof of Theorem~\ref{thm:sumprod} and Theorem~\ref{thm:sumprod_general}}
\label{sec:ProofSP}

We prove Theorem~\ref{thm:sumprod_general}, which covers Theorem~\ref{thm:sumprod} as a special case for $\alpha = 0$.
The required quantity is decomposed as
\begin{equation}
    \begin{aligned}
    \widetilde{R}_\alpha^{\mathcal S}
    &=
    \sum_{\mathbf m \in \mathcal G} \alpha^{ \sum_{l=0}^L \mathbbm{1}(m_l \notin \mathcal S)} R^{\mathbf m} 
    =
    \sum_{\mathbf m \in \mathcal G} \alpha^{ \sum_{l=0}^L \mathbbm{1}(m_l \notin \mathcal S)} \sum_{n=1}^{N^{(0)}} r_n^{0, \mathbf m} 
 =
                 \sum_{m_{0} \in \mathcal G}  \alpha^{\mathbbm{1}(m_{0} \notin \mathcal{S})}
         \sum_{n=1}^{N^{(0)}}
      \widetilde{  r}_n^{\;0, m_0, m_{1: L} \subseteq \mathcal G} , 
    \end{aligned}
\end{equation}
where 
\begin{align}
 \widetilde{ \mathbf r}^{\; 0, m_0, m_{1: L} \subseteq \mathcal G}
        =& \sum_{m_{1:L} \in \mathcal G} 
        \alpha^{\sum_{l=1}^L \mathbbm{1}(m_l \notin \mathcal S)}
         \mathbf T^{0, m_0, m_{1}} \mathbf T^{1, m_{1}, m_{2}}    \cdots   \mathbf T^{L-1, m_{L-1}, m_{L}} {\mathbf  r}^{\; L, m_{L}}
        \notag\\
        =& \sum_{m_{1} \in \mathcal G} \alpha^{\mathbbm{1}(m_{1} \notin \mathcal{S})} \mathbf T^{0, m_0, m_{1}} 
        \underbrace{\sum_{m_{2} \in \mathcal G} \alpha^{\mathbbm{1}(m_{2} \notin \mathcal{S})} \mathbf T^{1, m_{1}, m_{2}}  \quad  \cdots \quad \underbrace{\sum_{m_{L} \in \mathcal G} \alpha^{\mathbbm{1}(m_{L} \notin \mathcal{S})} \mathbf T^{L-1, m_{L-1}, m_{L}} { \mathbf  r}^{\; L, m_{L}}}_{= 
        \widetilde{\mathbf  r}^{\; L-1, m_{L-1}, m_{L} \subseteq \mathcal G}
        }
        }_{=
        \widetilde{ \mathbf  r}^{\; 1, m_{1}, m_{2:L} \subseteq \mathcal G}
        }.       
    \label{eq:GCN_GNN_LRP_rule_general1}
\end{align}
This decomposition gives the LRP rule \eqref{eq:GCN_GNN_LRP_rule_general} as a sum-product message passing
with the propagated relevance $\breve{\mathbf r}^{(l, m_l)} =  \widetilde{ \mathbf  r}^{\; l, m_{l}, m_{l+1:L} \subseteq \mathcal G}$, and proves Theorem~\ref{thm:sumprod_general}.
Noting that
\begin{align}
\widetilde{ \mathbf  r}^{\; l, m_{l}, m_{l+1:L} \subseteq \mathcal G}
=
{ \mathbf  r}^{\; l, m_{l}, m_{l+1:L} \subseteq \mathcal S}
\qquad \mbox{ for } \qquad \alpha = 0
\end{align}
proves Theorem~\ref{thm:sumprod}.
\QED

\section{Stardard LRP as Message Passing}
\label{sec:LRPasSumProduct}

Here, we derive the standard LRP rule as a sum-product message passing.
Consider a plain feed forward neural networks with $N^{(l)}$ neurons  at layer $l = 0, \ldots, L$.
The relevance of the (neuron-level) walk $\mathbf{n} = (n_0, \ldots, n_L)$ is then given as
\begin{equation}
    R^{\mathbf{n}} 
    = T^{0}_{n_{0},n_{1}}  T^{1}_{n_{1},n_{2}}  \quad  \cdots \quad  T^{L-1}_{n_{L-1},n_{L}} r_{n_{L}}^{L}
        =
    \left(    \prod_{l=0}^{L-1}
    T^{l}_{n_{l},n_{l+1}}
    \right)
    r_{n_{L}}^{L},
\end{equation}
where
$\mathbf r^{l} \in \mathbb{R}^{N^{(l)}}$ is such that $ r_n^{l}$ denotes the sum of relevances over all walks going through the neuron $n$ at layer $l$.
Collecting the relevances of all walks starting from the input node $n_0$ amounts to
\begin{equation}
    r_{n_0}^{0} = \sum_{n_{1}=1}^{N^{(1)}}
    \sum_{n_{2}=1}^{N^{(2)}}
    \cdots \sum_{n_{L}=1}^{N^{(L)}}
T^{0}_{n_{0},n_{1}}  T^{1}_{n_{1},n_{2}}  \quad  \cdots \quad  T^{L-1}_{n_{L-1},n_{L}} r_{n_{L}}^{L}.
\end{equation}
This summation can be decomposed as
\begin{equation}
    r_{n_0}^{0} = \sum_{n_{1}=1}^{N^{(1)}}
T^{0}_{n_{0},n_{1}}  
 \underbrace{ \sum_{n_{2}=1}^{N^{(2)}}
T^{1}_{n_{1},n_{2}}  \quad  \cdots \quad 
 \underbrace{  \sum_{n_{L}=1}^{N^{(L)}}
T^{L-1}_{n_{L-1},n_{L}} r_{n_{L}}^{L}}
_{= 
r_{n_{L-1}}^{L-1}
}
}_{= 
r_{n_{1}}^{1}
},
\end{equation}
giving a sum-product message passing:
\begin{equation}
 { r}_{n_l}^{l} = \sum_{n_{l+1}} T^{l}_{n_{l},n_{l+1}} 
   { r}_{n_{l+1}}^{l+1} 
\qquad \mbox{ or } \qquad
  {  \mathbf{r}}^{l} =  \mathbf{T}^{l}
  {  \mathbf{r}}^{l+1} .
\end{equation}
This coincides with the standard LRP \cite{bach2015pixel}.

\section{GNN-LRP as Message Passing}

\label{sec:ProofSPGNNLRP}

We can drive the original GNN-LRP rule as message passing, giving another proof of the following theorem:
\begin{theorem}
\citep{schnake2020higher}
\label{thm:sumprodGNNLRP}
  Assume that the GNN-LRP rule \eqref{eq:GCN_GNN_LRP_rule} is applied for $l = L-1, \ldots, 0$ with the initial message $\breve{\mathbf{r}}^{(L, m_L)} = \mathbf{r}^{L, m_L}$.  Then, 
  $\breve{\mathbf{r}}^{(l, m_l)}  = \mathbf{r}^{l, m_{l: L }}\; \forall l \in \{0, \ldots, L\} $, and 
  $R^{\mathbf m} =  \sum_{n=1}^{N^{(0)}}  r_n^{0, \mathbf m} $.
\end{theorem}
(Proof)
The required quantity is decomposed as
\begin{equation}
    \begin{aligned}
    {R}^{\mathbf m}
 =
     \sum_{n=1}^{N^{(0)}}
      {  r}_n^{0, \mathbf m} , 
    \end{aligned}
\end{equation}
where 
\begin{align}
 { \mathbf r}^{0,  \mathbf{m}}
        =&  \mathbf T^{0, m_0, m_{1}} 
        \underbrace{ \mathbf T^{1, m_{1}, m_{2}}  \quad  \cdots \quad \underbrace{\mathbf T^{L-1, m_{L-1}, m_{L}} { \mathbf  r}^{L, m_{L}}}_{= 
        {\mathbf  r}^{L-1,   m_{L-1:L} }
        }
        }_{= 
        { \mathbf  r}^{1,   m_{1:L} }
        }.       
    \label{eq:GCN_GNN_LRP_rule_general2}
\end{align}
This decomposition gives the LRP rule \eqref{eq:GCN_GNN_LRP_rule} 
as a sum-product message passing
with the propagated relevance $\breve{\mathbf r}^{(l, m_l)} = { \mathbf  r}^{ l,  m_{l:L} }$, and proves the theorem.
\QED

\section{Propagation Rules of Table \ref{table:GeneralLRPTable} for GCNs with the Notation in \citet{schnake2020higher}}
\label{appendix:schnake-notation}

To bring further intuition on the proposed propagation rules, we rewrite all propagation rules of Table \ref{table:GeneralLRPTable} in a notation similar to that of the original paper on GNN-LRP \cite{schnake2020higher}. Note that the defined notation here applies only in this appendix. We show the propagation rules specifically for the Graph Convolution Networks \cite{DBLP:conf/iclr/KipfW17}, with network connectivity encoded in the matrix $\Lambda$, weights at a given layer stored in a matrix $W$, and $h_J^a$ denoting the activation of neuron $a$ in node $J$. We denote by $\dots JKL \dots$ node indices in successive layers and jointly forming a walk. We denote by $a$ and $b$ two neurons indices (within their corresponding nodes) in successive layers. With this notation, the GCN equation for a given layer can be written as
$$
\forall_{K,b}:~~h_K^b = \max\Big(0,\sum_{J,a} \lambda_{JK} h_J^a w_{ab}\Big).
$$
Furthermore, we denote by $\mathcal{S}$ the set of nodes composing the subgraph of interest $\mathcal{S}$. It can also be interpreted as a coarse-graining of the member nodes into a new single node denoted by $\mathcal{S}$. We denote by $\sum_K$ the sum over all nodes in the given layer, and $\sum_b$ the sum over all neuron indices of a node in the given layer. The propagation rules can then be written as follows:
\begin{align*}
R_{J}^a &= \sum_{K,b} \frac{\lambda_{JK} h_J^a w_{ab}^\uparrow}{\sum_{J,a} \lambda_{JK} h_J^a w_{ab}^\uparrow} R_{K}^b, & \text{(LRP for general FFNN)}\\
R_{JKL\dots}^a &= \sum_b \frac{\lambda_{JK} h_J^a w_{ab}^\uparrow}{\sum_{J,a} \lambda_{JK} h_J^a w_{ab}^\uparrow} R_{KL\dots}^b, & \text{(GNN-LRP)}\\
R_{J\mathcal{S}\mathcal{S}\dots}^a &= \sum_{K \in \mathcal{S},b} \frac{\lambda_{JK} h_J^a w_{ab}^\uparrow}{\sum_{J,a} \lambda_{JK} h_J^a w_{ab}^\uparrow} R_{K\mathcal{S}\dots}^b, & \text{(sGNN-LRP)}\\
{R}_{J\mathcal{S}\mathcal{S}\dots}^{a,\alpha} &= \sum_{K,b} \alpha^{\mathbbm{1}(K \in \mathcal{S})}\frac{\lambda_{JK} h_J^a w_{ab}^\uparrow}{\sum_{J,a} \lambda_{JK} h_J^a w_{ab}^\uparrow} {R}_{K\mathcal{S}\dots}^{b,\alpha}.
& \text{(Generalized sGNN-LRP)}
\end{align*}

\section{Forward-hook Trick for Walk Relevance by GNN-LRP}
\label{FHDetail}

The forward-hook trick \cite{schnake2020higher, samek2021explaining} for computing the relevance of a walk works as follows.
We implement the forward combine step in Eq.\eqref{eq:GNN_agg_comb} as
\begin{equation}
    \begin{aligned}
        \mathbf P^{(l)} &\leftarrow \mathbf Z^{(l)} \mathbf W^{(l)\uparrow}, \\ 
        \mathbf Q^{(l)} &\leftarrow \mathbf P^{(l)} \odot [\sigma(\mathbf Z^{(l)}\mathbf W^{(l)}) \oslash \mathbf P^{(l)}]_{\texttt{cst.}}, \\ 
        \mathbf H^{(l)} &\leftarrow \mathbf Q^{(l)} \odot \mathbf M^{(l)} + [\mathbf Q^{(l)}]_{\texttt{cst.}} \odot(1-\mathbf M^{(l)}) ,
    \end{aligned}
    \label{eq:A.GCN_GNN_LRP_subgraph_rel_forward_hook_part1}
\end{equation}
where $\mathbf P^{(l)}, \mathbf M^{(l)} \in \mathbb R^{M \times N}$, and $\odot$ and $\oslash$ denote the entry-wise multiplication and division, respectively.
The operator $[\cdot]_{\texttt{cst.}}$ \emph{detaches} the quantity to which it applies from the gradient computation.
\citet{schnake2020higher} have shown that, if the \emph{mask} matrices $\{\mathbf M^{(l)}\}$ are set such that the column corresponding to the node $m_l$ specified by the walk is all-one vector, and the other columns are all-zero vectors, the relevance of a walk is obtained by
    \begin{align}
        R^{\mathbf  m} = \left<\texttt{Autograd}(y,\mathbf H^{(0)}), \mathbf M^{(0)} \mathbf H^{(0)}\right>,
\notag
        \end{align}
where $\left<\cdot, \cdot \right>$ denotes the Frobenius inner product.

\section{Proof of Theorem~\ref{thm:forward_hook} and Theorem~\ref{thm:forward_hook_general}}
\label{sec:ProofFH}

We prove Theorem~\ref{thm:forward_hook_general}, which covers Theorem~\ref{thm:forward_hook} as a special case for $\alpha = 0$.
The adapted forward computation of the model gives
\begin{equation}
    \begin{aligned}
        Z^{(l)}_{m_l,n_{l-1}} = & \sum_{m_{l-1} \in \mathcal G} \Lambda _{m_{l-1},m_{l}}H^{(l-1)}_{m_{l-1},n_{l-1}}, \\
        P^{(l)}_{m_{l},n_{l}} = & \sum_{n_{l-1}} Z^{(l)}_{m_l,n_{l-1}} W^{(l)\uparrow}_{n_{l-1},n_{l}}, \\
        Q^{(l)}_{m_{l},n_{l}} = & P^{(l)}_{m_{l},n_{l}} \left[ \frac{\rho (\sum_{n_{l-1}} Z^{(l)}_{m_{l},n_{l-1}} W^{(l)}_{n_{l-1},n_{l}})} {P^{(l)}_{m_{l},n_{l}}} \right]_{\texttt{cst.}}, \\
        H^{(l)}_{m_{l},n_{l}} = &  \alpha^{\mathbbm{1}(m_{l} \notin \mathcal{S})} Q^{(l)}_{m_{l},n_{l}} + (1-\alpha^{\mathbbm{1}(m_{l} \notin \mathcal{S})}) \left[Q^{(l)}_{m_{l},n_{l}}\right]_{\texttt{cst.}}.
    \end{aligned}
    \label{eq:GCN_GNN_LRP_fh1}
\end{equation}
Then the gradient is
\begin{equation}
    \begin{aligned}
        \frac{\partial H^{(l)}_{m_{l},n_{l}}}{\partial H^{(l-1)}_{m_{l-1},n_{l-1}}} = & 
                \alpha^{\mathbbm{1}(m_{l} \notin \mathcal{S})} \Lambda _{m_{l-1},m_{l}} W^{(l)\uparrow}_{n_{l-1},n_{l}} \frac{H^{(l)}_{m_{l},n_{l}}} {P^{(l)}_{m_{l},n_{l}}},
    \end{aligned}
    \label{eq:GCN_GNN_LRP_fh2}
\end{equation}
where we used $H^{(l)}_{m_{l},n_{l}} = \rho (\sum_{n_{l-1}} Z^{(l)}_{m_{l},n_{l-1}} W^{(l)}_{n_{l-1},n_{l}})$. 
By using 
the transition coefficient
\begin{equation}
    T_{n_{l-1}, n_{l}}^{l-1, m_{l-1}, m_l} = 
\frac{\Lambda_{m_{l-1}, m_l} H_{m_{l-1}, n_{l-1}}^{(l-1)} W^{(l-1)\uparrow}_{n_{l-1}, n_{l}} }{\sum_{m' ,n'} {\Lambda_{m', m_l} H_{m', n'}^{(l-1)} W^{(l-1)\uparrow}_{n', n_l } }}
    \label{eq:GCN_GNN_LRP_fh3}
\end{equation}
and Eq.\eqref{eq:GCN_GNN_LRP_fh1},
the gradient 
\eqref{eq:GCN_GNN_LRP_fh2}
can be written as
\begin{equation}
    \begin{aligned}
        \frac{\partial H^{(l)}_{m_{l},n_{l}}}{\partial H^{(l-1)}_{m_{l-1},n_{l-1}}} = & 
                \alpha^{\mathbbm{1}(m_{l} \notin \mathcal{S})} \underbrace{ \frac{\Lambda _{m_{l-1},m_{l}} W^{(l)\uparrow}_{n_{l-1},n_{l}}H^{(l-1)}_{m_{l-1},n_{l-1}}} {P^{(l)}_{m_{l},n_{l}}}}_{T_{n_{l-1}, n_{l}}^{l-1, m_{l-1}, m_l}} \frac{H^{(l)}_{m_{l},n_{l}}}{H^{(l-1)}_{m_{l-1},n_{l-1}}}\\
                = & \alpha^{\mathbbm{1}(m_{l} \notin \mathcal{S})} T_{n_{l-1}, n_{l}}^{l-1, m_{l-1}, m_l} \frac{H^{(l)}_{m_{l},n_{l}}}{H^{(l-1)}_{m_{l-1},n_{l-1}}}.
    \end{aligned}
    \label{eq:GCN_GNN_LRP_fh4}
\end{equation}
By applying the chain rule, we have the gradient of the output layer with respect to the input layer as
\begin{equation}
    \begin{aligned}
        \frac{\partial H^{(L)}_{m_{L},n_{L}}}{\partial H^{(0)}_{m_{0},n_{0}}} = & \sum_{m_{1},\cdots,m_{L-1}}\sum_{n_{1},\cdots,n_{L-1}} \frac{\partial H^{(L)}_{m_{L},n_{L}}}{\partial H^{(L-1)}_{m_{L-1},n_{L-1}}} \cdots \frac{\partial H^{(1)}_{M_{1},n_{1}}}{\partial H^{(0)}_{M_{0},n_{0}}}.
    \end{aligned}
    \label{eq:GCN_GNN_LRP_fh5}
\end{equation}
Substituting Eq.\eqref{eq:GCN_GNN_LRP_fh4} into Eq.\eqref{eq:GCN_GNN_LRP_fh5} gives
\begin{equation}
    \begin{aligned}
        \frac{\partial H^{(L)}_{m_{L},n_{L}}}{\partial H^{(0)}_{m_{0},n_{0}}} = &
        \sum_{m_{1},\cdots,m_{L-1}}\sum_{n_{1},\cdots,n_{L-1}} 
        \alpha^{ \sum_{l=1}^L \mathbbm{1}(m_l \notin \mathcal S)}
        \prod_{l=1}^{L} T_{n_{l-1}, n_{l}}^{l-1, m_{l-1}, m_l}
        \frac{H^{(L)}_{m_{L},n_{L}}}{H^{(0)}_{m_{0},n_{0}}}.
    \end{aligned}
    \label{eq:GCN_GNN_LRP_fh6}
\end{equation}

For the readout function, if we modify it according to the forward-hook trick described in Appendix B in \citet{samek2021explaining}, we can obtain the relevance of the $L$-th layer of GNN as 
\begin{align}
    r^{L, m_L}_{n_L} = H^{(L)}_{m_L, n_L} \frac{\partial y}{\partial H^{(L)}_{m_L, n_L}}.
\end{align}
Then, we have
\begin{equation}
    \begin{aligned}
        \frac{\partial y}{\partial H^{(0)}_{m_{0},n_{0}}} &= 
        \sum_{m_{1},\cdots,m_{L}}\sum_{n_{1},\cdots,n_{L}} 
        \alpha^{ \sum_{l=1}^L \mathbbm{1}(m_l \notin \mathcal S)}
        \prod_{l=1}^{L} T_{n_{l-1}, n_{l}}^{l-1, m_{l-1}, m_l}
        \frac{H^{(L)}_{m_{L},n_{L}}}{H^{(0)}_{m_{0},n_{0}}}\frac{\partial y}{\partial H^{(L)}_{m_{L},n_{L}}}\\
        &= 
        \sum_{m_{1},\cdots,m_{L}}\sum_{n_{1},\cdots,n_{L}} 
        \alpha^{ \sum_{l=1}^L \mathbbm{1}(m_l \notin \mathcal S)}
        \prod_{l=1}^{L} T_{n_{l-1}, n_{l}}^{l-1, m_{l-1}, m_l}
        \frac{H^{(L)}_{m_{L},n_{L}}}{H^{(0)}_{m_{0},n_{0}}}\frac{r^{L, m_L}_{n_L}}{H^{(L)}_{m_{L},n_{L}}}\\
        &= 
        \sum_{m_{1},\cdots,m_{L}}\sum_{n_{1},\cdots,n_{L}} 
        \alpha^{ \sum_{l=1}^L \mathbbm{1}(m_l \notin \mathcal S)}
        \prod_{l=1}^{L} T_{n_{l-1}, n_{l}}^{l-1, m_{l-1}, m_l}
        \frac{r^{L, m_L}_{n_L}}{H^{(0)}_{m_{0},n_{0}}}.
    \end{aligned}
\end{equation}
Multiplying the gradient with the masked initial activation and summing them up gives the final inner product:
\begin{equation}
    \begin{aligned}
        \sum_{m_{0},n_{0}} \alpha^{ \mathbbm{1}(m_0 \notin \mathcal S)} H^{(0)}_{m_{0},n_{0}} \frac{\partial y}{\partial H^{(0)}_{m_{0},n_{0}}} = &
        \sum_{m_{0},\cdots,m_{L}}\alpha^{ \sum_{l=0}^L \mathbbm{1}(m_l \notin \mathcal S)}\sum_{n_{0},\cdots,n_{L}} 
        \prod_{l=1}^{L} T_{n_{l-1}, n_{l}}^{l-1, m_{l-1}, m_l}
        r^{L, m_L}_{n_L} \\
    = &
        \sum_{\mathbf m}
        \alpha^{ \sum_{l=0}^L \mathbbm{1}(m_l \notin \mathcal S)}
        R^{\mathbf m} = \widetilde{R}^{\mathcal S}_\alpha,
    \end{aligned}
\end{equation}
which proves the theorem.
\QED

\section{Details of Datasets and GNN Models used in Experiments }
\label{sec:ModelDetails}
\subsection{BA2-Motif}

\begin{table*}[t]
    \caption{Statistics of the 5 datasets used in experiments.}
    \label{tab:stat_datasets}
    \vskip 0.15in
    \begin{center}
        \begin{small}
            \begin{sc}
                \begin{tabular}{lccccc}
                    \toprule
                     & BA-2motif & MUTAG & Mutagenicity & REDDIT-BINARY & Graph-SST2\\
                    \midrule
                    \# of Edges (avg) & 25.48 & 19.79 & 17.79 & 497.75 & 19.40 \\
                    \# of Nodes (avg) & 25.00 & 17.93 & 16.90 & 429.63 & 10.20 \\
                    \# of Graphs & 1000 & 188 & 4337 & 2000 & 70042 \\
                    
                    \bottomrule
                \end{tabular}
            \end{sc}
        \end{small}
    \end{center}
    \vskip -0.1in
\end{table*}

BA-2motif \cite{luo2020parameterized} is a synthetic dataset of graphs that can be classified into two classes according to the different motifs. For each sample graph, a base graph is generated by the Barab\'asi-Albert (BA) model, and then one of two motifs is connected to it. Because this dataset is synthesized, the ground truth about which nodes build the motif is available.

We trained GIN models with 2,3,4,5,6,7 layers, and all models has the same GIN block, which is a 2-layer MLP. The input feature dimension is 1 and the output feature dimension in the MLP blocks $N^{(l)} = 20, \  \forall l = 1, \cdots, L-1$. The activation function is ReLU. We employed the SGD optimizer with a decreasing learning rate $\gamma = 0.00001 / (1.0 + (\texttt{epoch} / \texttt{epochs})$ for 10000, 10000, 5000, 1000, 1000, 1000 epochs, respectively. We downloaded the dataset from the repository of \citet{schnake2020higher}, and the dataset includes 1000 samples with the first 500 samples from positive class and the last 500 samples from negative class. Because all samples are randomly generated, it is unnecessary to sample randomly and we used the 0-400 and 500-900 as training dataset and the rest as testing dataset. The test accuracy of the three models are 98\%, 99.50\%, 100\%, 100\%, 100\%, 100\%, respectively.

\subsection{MUTAG}

MUTAG \cite{debnath1991structure} is a datasets of molecules. Every sample graph includes atoms as nodes and chemical links as edges. The molecules are labeled as mutagenic or non-mutagenic. 
The node feature is the node type (atom), which is represented as a one-hot vector.

The 3-layered GIN model has in all layers a 2-layer MLP as the GIN blocks. The input feature dimension is 7, which is one-hot vectors denoting different atoms. The output dimensions in the MLP blocks are 128. We used 108 samples with half positive and half negative as training dataset, and the rest samples build the testing dataset. We trained the model with SGD optimizer for 1500 epochs, and the learning rate $\gamma = 0.0005 / (1.0 + (\texttt{epoch} / \texttt{epochs})$. The test accuracy is 85.00\%.

\subsection{Mutagenicity}\label{app:mutagenicity}

Mutagenicity \cite{doi:10.1021/jm040835a} is another dataset for chemical molecules, which is larger than MUTAG and contains much more variety of mutagenic molecules with  different types of toxic groups.

The model setting is the same as for MUTAG, except the input feature dimension increased to 13 as there are more atoms in this dataset. We used 3096 samples with half positive and half negative as training dataset, and the rest as testing dataset. We trained the model with Adam optimizer for 25 epochs, and the initial learning rate $\gamma = 0.00005$. The test accuracy is 83.16\%.

\subsection{REDDIT-BINARY}

REDDIT-BINARY \cite{DBLP:conf/kdd/YanardagV15} is a social network dataset, and each graphs stands for a community, with nodes being users and edges denoting that there is at least one response to the comments between the two users. The graphs are classified into two classes according to which kind of community the users build, i.e., question/answer-based community or discussion-based community. 
The dataset contains large graphs ($> 400$ nodes in average), and no node feature is provided.

Our model for this dataset is 5-layer GIN, with 2-layer MLP as GIN blocks. The input feature dimension is 1 and the output dimensions of the GIN blocks are 64. The training dataset has 1600 samples of half positive and half negative, and the rest 400 samples build the testing dataset. We trained the model with Adam optimizer for 500 epochs and the initial learning rate $\gamma = 0.00005$. The test accuracy is 82.50\%.

\subsection{Graph-SST2}

Graph-SST2 \cite{yuan2020explainability} is a dataset of texts labeled in two sentiment classes. The text are transformed into parse tree graphs with 768-dimensional embedded vectors of the words for the initial node features.

The model is built with a node feature embedding part and a following 3-layer GCN. The input feature dimension is 768, and in the middle layer of GCN the output dimension is 20. We downloaded the dataset from \citet{yuan2020explainability} and used their dataset split. We trained the model with Adam optimizer for 50 epochs, and the initial learning rate $\gamma = 0.0001$. The test accuracy is 89.40\%.

\section{Details of sGNN-LRP Implementation
}
\label{sec:method_algos}

Algorithm \ref{alg:mp_fw_subgraph_rel_alphazero}
and 
Algorithm
\ref{alg:mp_fw_subgraph_rel}, respectively, summarize the procedures of sGNN-LRP (with the Forward-hook trick) for the original subgraph attribution ($\alpha = 0$) and for the generalized subgraph attribution ($\alpha \in (0, 1]$).

\begin{algorithm}[tb]
    \caption{sGNN-LRP, $\alpha=0$}
    \label{alg:mp_fw_subgraph_rel_alphazero}
\begin{algorithmic}
    \STATE {\bfseries Input:} graph $\mathcal G$, subgraph $\mathcal S$, \# of model layers $L$.
    \STATE {\bfseries Output:} subgraph relevance score $R^{\mathcal S}$
    \STATE Initialize mask matrix $\mathbf M_\mathcal S$, such that $\mathbf M_\mathcal S$ is valued $1$ on the indices of nodes inside $\mathcal S$ and $0$ else.
    \FOR{$l=1$ {\bfseries to} $L$}
        \STATE $\mathbf Z^{(l)} \leftarrow \mathbf \Lambda \mathbf H^{(l-1)} $
        \STATE $\mathbf P^{(l)} \leftarrow \mathbf Z^{(l)} \mathbf W^{(l)\uparrow} $
        \STATE $\mathbf Q^{(l)} \leftarrow \mathbf P^{(l)} \odot [\rho(\mathbf Z^{(l)}\mathbf W^{(l)}) \oslash \mathbf P^{(l)}]_{cst.} $
        \STATE $\mathbf H^{(l)} \leftarrow \mathbf Q^{(l)} \odot \mathbf M_\mathcal S + [\mathbf Q^{(l)}]_{cst.} \odot(1-\mathbf M_\mathcal S)$
    \ENDFOR
    \STATE $y \leftarrow \text{readout}(\mathbf H^{(L)})$ with readout function modified according to LRP forward-hook trick.
    
    \STATE $R^{\mathcal S} = \sum_{m_{0} \in \mathcal S} \left<\text{Autograd}(y,\mathbf H^{(0)}_{m_{0}}), \mathbf H^{(0)}_{m_{0}}\right>$
\end{algorithmic}
\end{algorithm}

\begin{algorithm}[tb]
    \caption{sGNN-LRP, $\alpha \in (0,1]$}
    \label{alg:mp_fw_subgraph_rel}
\begin{algorithmic}
    \STATE {\bfseries Input:} graph $\mathcal G$, subgraph $\mathcal S$, discount factor $\alpha$, \# of model layers $L$.
    \STATE {\bfseries Output:} subgraph relevance score $R^{\mathcal S}$
    \STATE Initialize mask vector $\mathbf M^\mathcal S_\alpha$, such that $\mathbf M^\mathcal S_\alpha$ is valued $1$ on the indices of nodes inside $\mathcal S$ and $\alpha$ else.
    \FOR{$l=1$ {\bfseries to} $L$}
        \STATE $\mathbf Z^{(l)} \leftarrow \mathbf \Lambda \mathbf H^{(l-1)} $
        \STATE $\mathbf P^{(l)} \leftarrow \mathbf Z^{(l)} \mathbf W^{(l)\uparrow} $
        \STATE $\mathbf Q^{(l)} \leftarrow \mathbf P^{(l)} \odot [\rho(\mathbf Z^{(l)}\mathbf W^{(l)}) \oslash \mathbf P^{(l)}]_{cst.} $
        \STATE $\mathbf H^{(l)} \leftarrow \mathbf Q^{(l)} \odot \mathbf M^\mathcal S_\alpha + [\mathbf Q^{(l)}]_{cst.} \odot(1-\mathbf M^\mathcal S_\alpha)$
    \ENDFOR
    \STATE $y \leftarrow \text{readout}(\mathbf H^{(L)})$ with readout function modified according to LRP forward-hook trick.
    
    \STATE $R_1 = \sum_{m_{0} \in \mathcal G} \left<\text{Autograd}(y,\mathbf H^{(0)}_{m_{0}}), \mathbf H^{(0)}_{m_{0}}\right>$
    
    \STATE Set mask vector $\mathbf M^\mathcal S_\alpha$, such that $\mathbf M^\mathcal S_\alpha$ is valued $0$ on the indices of nodes inside $\mathcal S$ and $\alpha$ else.
    \FOR{$l=1$ {\bfseries to} $L$}
        \STATE $\mathbf Z^{(l)} \leftarrow \mathbf \Lambda \mathbf H^{(l-1)} $
        \STATE $\mathbf P^{(l)} \leftarrow \mathbf Z^{(l)} \mathbf W^{(l)\uparrow} $
        \STATE $\mathbf Q^{(l)} \leftarrow \mathbf P^{(l)} \odot [\rho(\mathbf Z^{(l)}\mathbf W^{(l)}) \oslash \mathbf P^{(l)}]_{cst.} $
        \STATE $\mathbf H^{(l)} \leftarrow \mathbf Q^{(l)} \odot \mathbf M^\mathcal S_\alpha + [\mathbf Q^{(l)}]_{cst.} \odot(1-\mathbf M^\mathcal S_\alpha)$
    \ENDFOR
    \STATE $y \leftarrow \text{readout}(\mathbf H^{(L)})$ with readout function modified according to LRP forward-hook trick.
    
    \STATE $R_2 = \sum_{m_{0} \in \mathcal G} \left<\text{Autograd}(y,\mathbf H^{(0)}_{m_{0}}), \mathbf H^{(0)}_{m_{0}}\right>$
    
    \STATE $R^{\mathcal S} = R_1 - R_2$
\end{algorithmic}
\end{algorithm}

\section{Details of the Model Activation and Pruning Experiments}
\label{sec:detail_acti_prun}

Algorithm \ref{alg:model_activation} and Algorithm \ref{alg:model_pruning},
respectively,
describe the detailed procedures of model activation and pruning experiments (node ordering and its evaluation).

\begin{algorithm}[tb]
    \caption{Model Activation Task}
    \label{alg:model_activation}
\begin{algorithmic}
    \STATE {\bfseries Input:} GNN model $f(\cdot)$, input graph $\mathcal G$
    \STATE {\bfseries Output:} \texttt{AUAC}
    \STATE Find the node sequence $\left[m^{(1)},...,m^{(M)}\right]$ such that $\sum_{i=1}^{M} R^{\{m^{(1)},...,m^{(i)}\}}$ is maximized.
    \STATE Initialize $\texttt{AUAC} = 0$.
    \STATE $\mathcal S = \varnothing $
    \FOR{$ i = 1, \cdots, |\mathcal G| $}
        \STATE $\mathcal S = \mathcal S\cup \{m^{(i)}\}$
        \STATE $\texttt{AUAC} = \texttt{AUAC} + f(\mathcal S)$
    \ENDFOR
    \STATE $\texttt{AUAC} = \texttt{AUAC} / |\mathcal G|$
\end{algorithmic}
\end{algorithm}

\begin{algorithm}[tb]
    \caption{Model Pruning Task}
    \label{alg:model_pruning}
\begin{algorithmic}
    \STATE {\bfseries Input:} GNN model $f(\cdot)$, input graph $\mathcal G$
    \STATE {\bfseries Output:} \texttt{AUPC}
    \STATE Find the node sequence $\left[m^{(1)},...,m^{(M)}\right]$ such that $\sum_{i=1}^{M} |R^{\mathcal G \backslash \{m^{(1)},...,m^{(i)}\}}-R^{\mathcal G}|$ is minimized.
    \STATE Initialize $\texttt{AUPC} = 0$.
    \STATE $\mathcal S = \varnothing $
    \FOR{$ i = 1, \cdots, M $}
        \STATE $\mathcal S = \mathcal S\cup \{m^{(i)}\}$
        \STATE $\texttt{AUPC} = \texttt{AUPC} + |f(\mathcal G \backslash \mathcal S)-f(\mathcal G)|$
    \ENDFOR
    \STATE $\texttt{AUPC} = \texttt{AUPC} / |\mathcal G|$
\end{algorithmic}
\end{algorithm}

\section{Subgraph Attribution using GNNExplainer}
\label{sec:gnn_alpha}

In Section~\ref{sec:APExperiment}, we applied the GNNExplainer to compute the subgraph relevance according to the definition of generalized subgraph relevance definition \eqref{eq:subgraph_rel_def_2}. 
GNN-Explainer attributes the edges, and 
we consider an edge as a one-step walk which contains only two nodes.
According to \eqref{eq:subgraph_rel_def_2}, the subgraph relevance is the sum of all edges that have at least one node inside the subgraph, with the partly-outside edge (one node inside and one node outside the subgraph) deweighted with $\alpha$, i.e.,
\begin{equation}
    R_{\mathrm{GNN-Exp}}^{\mathcal S} = \sum_{\mathbf m \in \mathcal G} g_\alpha^{\mathcal S}(\mathbf m) R_{\mathrm{GNN-Exp}}^{\mathbf m}, \qquad  
    g_\alpha^{\mathcal S}(\mathbf m) = \begin{cases}
                0 & \text{ if } m_1 \notin \mathcal S \land  m_2 \notin \mathcal S ,\\
       \alpha & \text{ if } m_1 \notin \mathcal S \lor  m_2 \notin \mathcal S ,\\
       1 & \text{ otherwise},\\
            \end{cases}
\end{equation}
where $\mathbf m \in \{1, \ldots, M\}^2$ denotes edges between nodes $m_1$ and $m_2$.

\end{document}